\PassOptionsToPackage{dvipsnames,svgnames,table}{xcolor}
\documentclass[11pt]{article}

\usepackage{tcolorbox}
\usepackage{CJKutf8}
\usepackage{times,latexsym}
\usepackage{url}
\usepackage[T1]{fontenc}
\usepackage{lineno}
\usepackage{lineno}
\usepackage{float}

\usepackage{ACL2023}

\usepackage{CJKutf8}

\usepackage{times}
\usepackage{latexsym}
\usepackage{booktabs}
\usepackage{siunitx} 
\usepackage{tabularx}
\usepackage{rotating}
\usepackage{subcaption}
\usepackage{array}
\usepackage{makecell}
\usepackage{multirow}
\usepackage{caption} 
\usepackage{booktabs}
\usepackage{enumitem}
\usepackage{multicol}
\usepackage{cuted}
\usepackage{changepage}
\usepackage[utf8]{inputenc}
\usepackage{tcolorbox}
\usepackage{listings}
\usepackage[table,dvipsnames,svgnames,x11names]{xcolor}
\usepackage{hyperref}
\hypersetup{
  pdfauthor={},
  pdftitle={},
  pdfcreator={},
  pdfproducer={}
}
\usepackage{tabularx,booktabs}
\newcolumntype{L}{>{\raggedright\arraybackslash}X}
\newcolumntype{R}{>{\raggedright\arraybackslash}X}

\usepackage{xspace,mfirstuc,tabulary}

\newif\iftaclinstructions
\taclinstructionsfalse 
\iftaclinstructions

\newcommand{\instr}
\fi

\newcommand{\korean}[1]{{\begin{CJK*}{UTF8}{mj}#1\end{CJK*}}}
\usepackage[T1]{fontenc}
\usepackage{graphicx} 
\DeclareGraphicsExtensions{.pdf,.png,.jpg}

\usepackage[utf8]{inputenc}

\usepackage{microtype}

\usepackage{inconsolata}
\usepackage{rotating}
\usepackage{stfloats} 
\usepackage{placeins}   
\newtcolorbox{instructionsbox}[1][]{
  colframe=cyan!75!black,    
  colback=green!5!white,     
  coltitle=black,            
  title=#1,                  
  rounded corners,           
  boxrule=0.5mm,             
  boxsep=5pt,                
  toptitle=1mm,              
  bottomtitle=1mm,           
  left=0pt,                 
  right=0pt,                
  top=0pt,                   
  bottom=0pt,                
  fonttitle=\bfseries        
}

\title{A 2-step Framework for Automated Literary Translation Evaluation: Its Promises and Pitfalls}

\author{
  \textbf{Sheikh Shafayat} \quad
  \textbf{Dongkeun Yoon} \quad
  \textbf{Woori Jang} \quad
  \textbf{Jiwoo Choi} \\
  \textbf{Alice Oh} \quad
  \textbf{Seohyon Jung} \\
    KAIST \\
  \texttt{\{sheikh.shafayat,seohyon.jung\}@kaist.ac.kr}
}

\begin{document}
\maketitle
\begin{abstract}
In this work, we propose and test a two-step framework for evaluating literary machine translation, focusing on English to Korean. The framework combines RULER, a rubric-based evaluation tailored to language-specific features, with VERSE, a language-agnostic question–answering module. Our results show that it provides fine-grained, interpretable metrics suited for literary translation and achieves stronger correlation with human judgment than traditional machine translation metrics. However, it still falls short of inter-human agreement, particularly in culturally sensitive areas such as Korean honorifics. We also find that LLMs tend to favor translations produced by other LLMs. These findings highlight both the potential of our framework and the need for culturally aware evaluation methods to ensure accurate and sensitive translation of literary works. Code\footnote{\url{https://github.com/sheikhshafayat/ruler-verse}} and
dataset\footnote{\url{https://huggingface.co/datasets/skshafayat/two-step-lit-eval}}
are publicly available.

\end{abstract}

\section{Introduction}

Recent studies demonstrate that large language models (LLMs) show strong performance in machine translation \cite{zhang-etal-2023-machine, jiao2023chatgpt}. In fact, properties like steerability, context awareness, background knowledge, and context learning capabilities make LLMs suitable for grounded creative tasks like literary translation \cite{karpinska2023large}, which is considered the last frontier of machine translation \cite{klemin2024last, wang2023findings}. 

While new approaches such as agentic systems are being utilized for literary translation \cite{wu2024perhaps, he2024exploring}, evaluation remains challenging because automated metrics such as BLEU \cite{papineni2002bleu}, BLEURT \cite{sellam2020bleurt} are not capable of capturing the nuance of literary translation \cite{thai2022exploring}. 

Yet, for the agentic translation systems to work at scale, we need an automatic evaluation method first, which will give signal to the LLM agents and pipeline developer whether the translation meets the required criteria or not since human evaluation in each iteration is costly. Recently, however, pretrained LLMs have become more common in evaluating machine translation \cite{kocmi2023gemba, freitag2023results}, in lieu of and in collaboration with human evaluation.

\begin{figure}[t]
    \centering

    \includegraphics[width=0.48\textwidth]{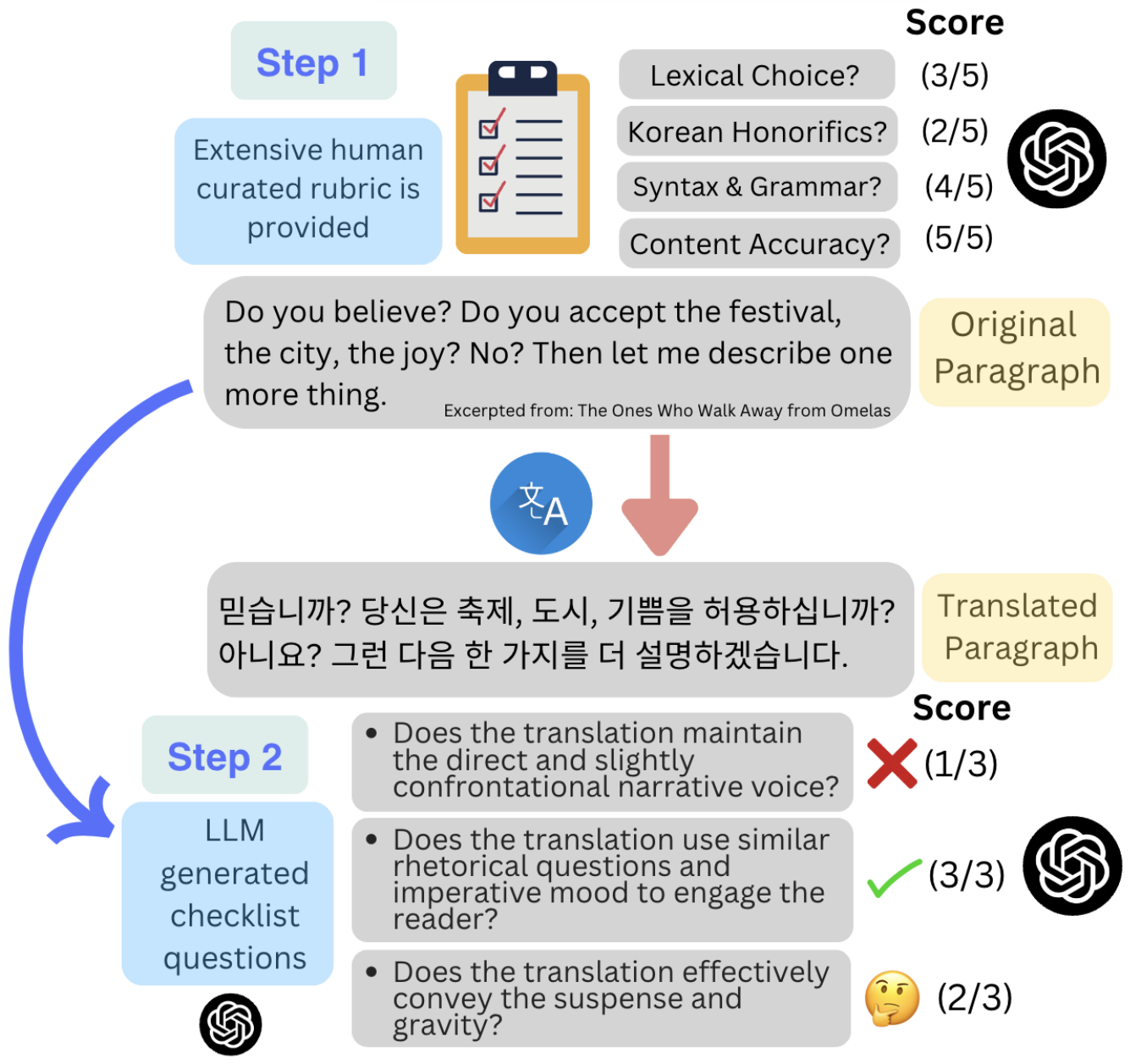}
    \caption{\small{The overview of our proposed framework: we evaluate translation of literary works in two stages. In the first step, we focus on \textit{semantic and linguistic} features of the translation using a human-generated rubric. In the second step, we verify how much of the \textit{literary qualities} have been captured through the translation by two-agent question answering, where one LLM generates a list of criteria and the other LLM verifies whether those criteria have been satisfied.}}
    \label{fig:concept}
\end{figure}

In this work, we propose a two-stage framework for literary translation evaluation. In the first stage, we evaluate literary translations with expert-curated Direct Assessment Multidimensional Quality Metrics (MQM) \cite{mariana2014multidimensional}, while in the second stage, we evaluate the translation using story-specific criteria generated by LLMs for a successful \textit{literary} translation. A verifier LLM then judges whether the machine translation meets the criteria.


Our analysis demonstrates that current LLMs approximate human preference in English–Korean literary translation on some dimensions, but they consistently underperform in culturally specific areas such as honorifics. This highlights both the potential of automated evaluation and its limits in capturing cultural features crucial for assessing literary translations. 


\section{Background}
\paragraph{Literary Translation}
Building on James S. Holmes's 1970s' foundational theories of literary translation, scholars like Susan Bassett, André LeFevere, and Lawrence Venuti have led the “cultural turn” in translation studies, which emphasizes the importance of cultural and historical contexts in literary translation \cite{snell2006turns}. Literary translation has a special place within broader translation studies for its focus on the aesthetic and affective use of language as well as its engagement with fictional content. Some of the persisting concerns in literary translation include the tension between literal (word-for-word) and free or adaptive (sense-for-sense) \cite{munday2022introducing}, the visibility of the translator \cite{venuti2017translator}, and the social role of the translator \cite{blakesley2018sociologies}. More recently, scholars including Kolb and Macken have examined the potential of the integration of technology and machine translation in literary translation processes \cite{kolb2023bit, macken2022literary}. Our work presents a pipeline incorporating the cultural and aesthetic aspects of literary translation into the evaluation of machine literary translation.

\paragraph{Machine Translation Evaluation}
String-based metrics like BLEU \cite{papineni2002bleu} have been widely used in machine translation evaluation. Similarly, more recent metrics such as BLEURT \cite{sellam2020bleurt}, COMET \cite{rei2020comet}, and XCOMET \cite{colombo2023xcomet} leverage BERT-style encoder language models \cite{devlin-etal-2019-bert} fine-tuned on translation preference dataset. 
However, previous work has shown that these metrics are not suitable for literary translation \cite{thai2022exploring}. Departing from earlier trends, \citet{kocmi2023gemba} employed the cutting-edge LLM GPT-4 \cite{openai2023gpt} to assess machine translation, where their approach, GEMBA-MQM, set a new benchmark for translation quality estimation at WMT 2023 \cite{freitag2023results}. Recent studies from \citet{wu2024perhaps} and \citet{chen2024benchmarkingllmstranslatingclassical} also use GPT-4 as a judge for literary translation for web novels and Chinese poetry, respectively.

We take a step back and conduct a comprehensive study on the LLMs' capabilities of fine-grained evaluation for \textit{literary} translation. A closely related work, \cite{zhang2024goodllmsliterarytranslation}, provide a broad human-study benchmark (LITEVAL-CORPUS) highlighting the inadequacy of MQM for literature and the limits of current automatic metrics.

\paragraph{LLM-as-a-Judge} 
LLM-as-a-judge has emerged as one dominant paradigm in evaluating model outputs \cite{zheng2023judging, liu2023g,dubois2023alpacafarm, chan2023chateval, li2023generative, kim2023prometheus}. While LLM judges are known to be on par with human judgment in generic tasks in MT-Bench and Chatbot Arena \cite{zheng2023judging}, several works have highlighted their shortcomings, such as biases in order of the choices \cite{wang2023large}, length of the response \cite{saito2023verbosity}, or language of the response \cite{son2024mmevalmultilingualmetaevaluationbenchmark}.

Motivated by earlier research on the effectiveness and limitations of LLM-as-a-judge, we aim to highlight the potential and perils of using LLMs as evaluators within the specific context of literary translation.
\paragraph{Question answering based self-evaluation} 
Our proposed question-answering-based translation evaluation method relates closely to automatic metrics like TIFA \cite{hu2023tifa}, VPEval \cite{cho2024visual}, and DSG \cite{cho2023davidsonian} for measuring the consistency of text-to-image generation models such as DALL-E \cite{pmlr-v139-ramesh21a} and Stable Diffusion \cite{podell2024sdxl}. In this line of work, an LLM first generates a list of verification questions from the given user's prompt, for example ``is there a car?" ``is the car blue?" and then passes the questions to a vision language model such as BLIP \cite{li2022blip} which answers the verification questions. This approach also has a resemblance to how previous works have measured faithfulness in summarization \cite{kim2024fables, wang2020asking} and long-form factuality evaluation \cite{min2023factscore}. Another work has \cite{krubinski-etal-2021-just} explored question-answering paradigms for machine translation evaluation before. In this work, we extend this paradigm to do fine-grained evaluation of \textit{literary} translation.


\section{Dataset, Metrics \& Annotators}
\paragraph{Dataset} We collect 15 short stories translated from English to Korean, representing a diverse range of genres and styles. We then manually align the stories paragraph-wise.  
The dataset comprises 725 parallel aligned paragraphs across stories, averaging 48.3 paragraphs per story with a standard deviation of 28.9. Each paragraph contains an average of 6.8 sentences, with a standard deviation of 6.9, in English. A full list of the stories and additional details can be found in Table \ref{tab:dataset} in Appendix \ref{app: dataset}.


 \paragraph{Metrics} We use expert human evaluation to verify the efficacy and viability of our proposed framework. In all steps, we use Kendall's Tau and Spearman correlation to measure the correlation with human judgment. We also report the mean squared error (MSE) with the scores provided by models with human annotators' scores. 

 \paragraph{Human Annotators} Three literary experts with extensive experience in both English and Korean literature develop and validate the framework presented in Section \ref{sec:approach}. They also carry out the annotations in Section \ref{sec:model_human_preference} to validate the framework.
\section{Two-Step Literary Translation Evaluation Framework}
\label{sec:approach}
Even experts continue to disagree on what constitutes a good literary translation. The nuanced interplay of language and culture in literature often produces multiple, equally valid translations \cite{sager1998distinguishes}. Moreover, translators sometimes have to make the difficult decision to deviate from the original text to culturally adapt the material so that the essence of the work is conveyed. A good literary translation often diverges from a faithful, word-to-word rendering. Evaluating literary translation thus demands unique criteria and standards compared to other machine translation tasks. 

In this work, we propose a two-step framework for evaluating literary translation (see Figure \ref{fig:concept}):
\paragraph{\textsc{Step 1}} 
\textbf{RULER:} \textbf{RU}bric-based \textbf{L}iterary \textbf{E}valuation \textbf{R}ating is a Likert scale of 1 to 5 with four criteria, which resemble Multidimensional Quality Metric (MQM) \cite{mariana2014multidimensional}, for evaluating translations based on rubrics. 
\paragraph{\textsc{Step 2}} 
\textbf{VERSE:} A \textbf{VER}ification-based \textbf{S}tory specific \textbf{E}valuation that asks \textit{story-specific literary questions} that consider the distinct characteristics needed in successful translation of literary works.

\subsection{\textsc{Step 1}: RULER}
\label{subsec: step 1}
We first conduct a small-scale qualitative evaluation of machine-translated literary texts to identify the capabilities that evaluation models should possess, with a particular focus on the types of errors they should detect. For this, we select five stories (marked dark gray in Table \ref{tab:dataset}) and translate them using various large and small language models, as well as commercially available translation services, like Google Translate, to get a diverse set of translations. We observe that LLMs show inconsistent quality in doing literary translation. Larger models like GPT-4o or Claude Sonnet are more likely to produce remarkably sophisticated translations, while smaller ones often make unexpected errors, even with relatively simple passages. Our analysis identifies four major types of errors machine translation systems make when translating literary works, some of which are especially pronounced in English-to-Korean translations: word/phrase/expression-level issues, context-related issues, issues relevant to honorifics, and form-related issues (see examples in Table \ref{table:common_mistakes}). We provide a detailed explanation of these categories with concrete examples in Appendix \ref{appendix:common_issues}.

\begin{figure*}[t]
  \centering
  \includegraphics[width=0.98\textwidth]{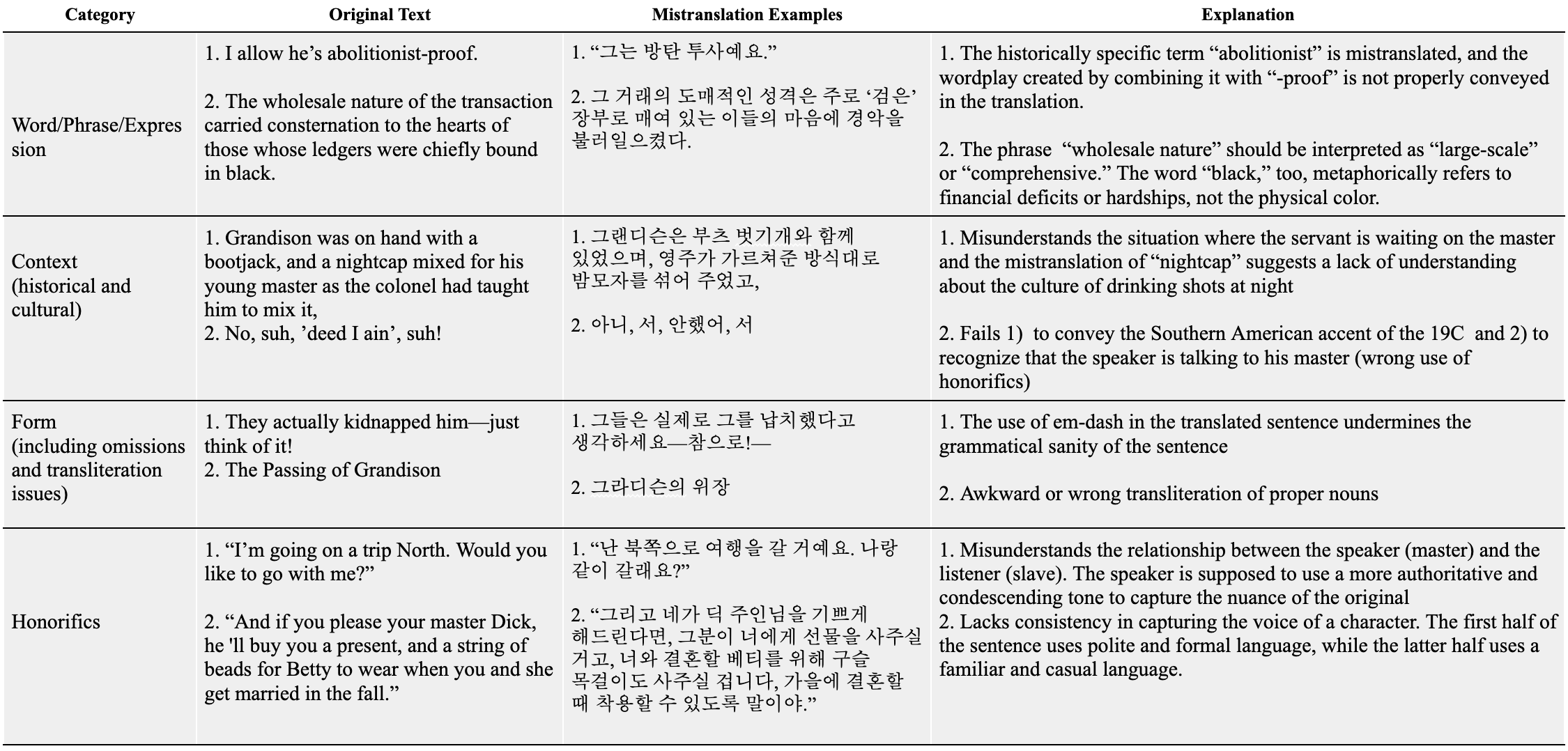}
  \captionof{table}{\small{Examples of LLM mistranslations for each category. Examples are from Chesnutt’s “The Passing of Grandison”(1899), a short story on the social construction of racial dynamics.}}
  \label{table:common_mistakes}
\end{figure*}

From this initial qualitative evaluation of translations, we identify four categories that are critical to the foundational quality of English-to-Korean literary translations (listed below). We then develop a fine-grained rubric to systematically assess translations, focusing on these categories designed to detect common mistake patterns in literary translation.
\begin{itemize}
    \item \textbf{Lexical Choice} Proper use of idiomatic expressions and general word choices that demonstrate fluency in the target language.
    \item \textbf{Proper Use of Honorifics in Dialogues} Selection of verb endings that accurately reflect the status of and relationships among the participants in a conversation, ensuring natural-sounding dialogue in Korean. Note that this applies only to paragraphs containing dialogue or other expressions of register in the description.
    \item \textbf{Syntax and Grammar} Grammatical accuracy of sentences including proper translation of embedded clauses, conjunctions, sentence connectors, internal logic of sentences, and punctuations, which ultimately makes the translation read naturally in Korean.
    \item \textbf{Content Accuracy} Preservation of the text’s meaning, with particular emphasis on proper and comprehensive translation of literary expressions in the original text. While accurate translation might involve some additions or omissions for clarity, it should avoid altering or omitting substantial words or phrases that are essential to the content and context of the original.
\end{itemize}
The detailed rubric for each criterion will be released with our code and data after acceptance.

\subsection{\textsc{Step 2}: VERSE}
\label{subsec: step 2 details}


Eliminating common errors is insufficient for achieving high-quality literary translation. A successful translation must also maintain the essential and unique literary qualities of the original work, which can vary significantly between different texts. To capture such nuances, we propose VERSE, a question-answering-based evaluation framework. We prompt a high-performing LLM (in our case GPT-4o \cite{openai2024gpt4o}) to generate a list of verification questions that must be satisfied in a good translation. We instruct the question-generating model to focus on the literary aspects of the passage, with a brief summary of the story provided for context. Figure \ref{fig:concept} provides examples of the questions generated by the LLM.


Another LLM then evaluates the machine translation based on these questions. The grading was done on a 1 to 3 scale, where 1 indicates the criteria in the question have not been satisfied at all, 2 indicates partial satisfaction, and 3 indicates full satisfaction of the criteria.

One expert human inspected 50 randomly sampled generated questions and found 44 of them to be ``very helpful". The six remaining questions, while relevant, were either too generic or unanswerable based on the provided context alone.

\begin{figure}[ht]
    \includegraphics[width=0.4\textwidth]{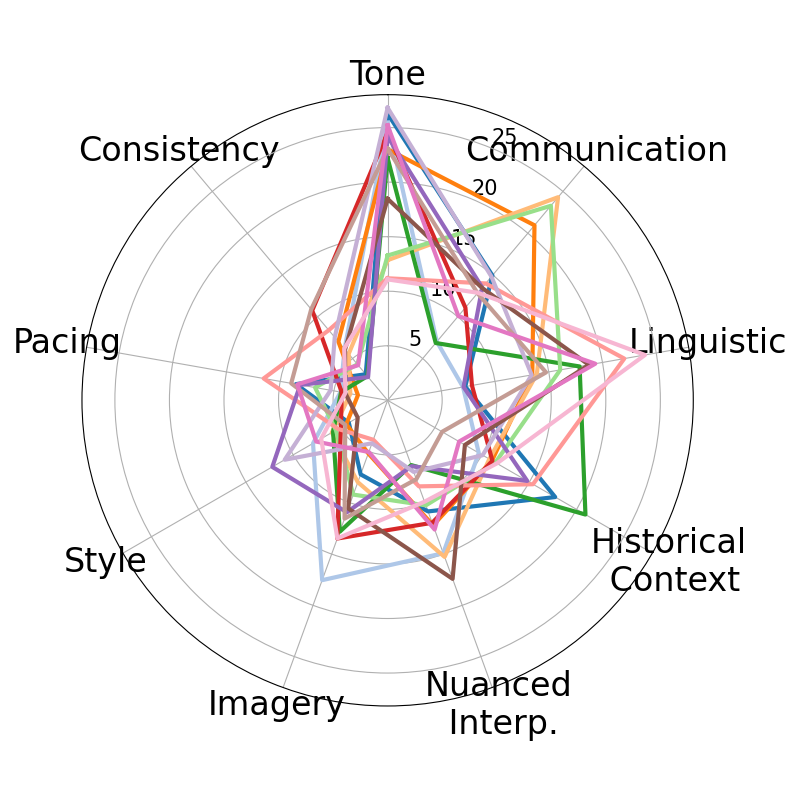}
    
    \caption{\small{Different stories require different kinds of attention to ensure that the literary essence of the work is preserved in the translation. The categories were found after classifying them in zero shot manner. See Section \ref{subsec: step 2 details}. The radial axis represents the percentage of questions belonging to that category for that particular story.}}
    \label{fig:story-specific}
\end{figure}

\begin{figure}[t]
    \includegraphics[width=0.48\textwidth]{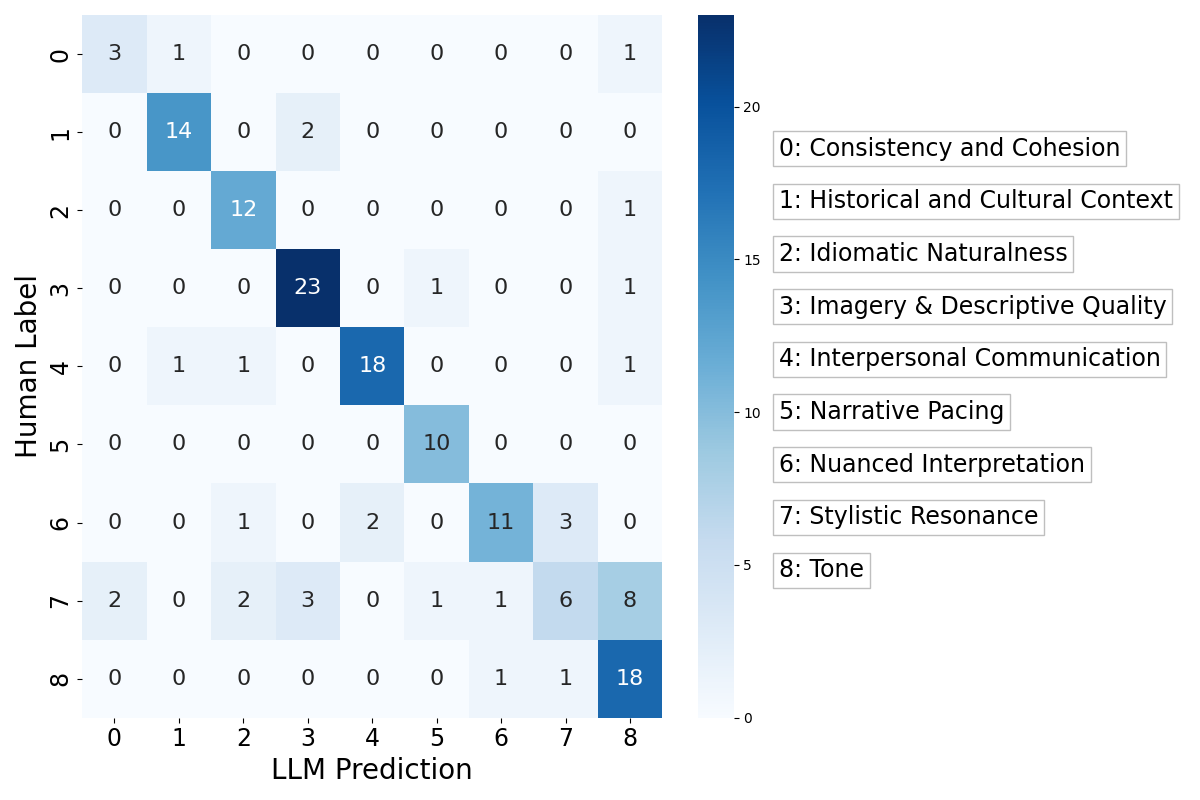}
    \caption{\small{Confusion matrix of the zero-shot question classification. The category names are slightly abridged to fit in the figure. Notice that most apparent misclassifications are usually in a very closely related class and human inspection revealed that in most cases the misclassified class can also be considered valid.}}
    \label{fig:confusion-matrix}
\end{figure}

\paragraph{Clustering LLM Generated Questions} 
To provide better interpretability for the LLM generated questions in VERSE, we classify the questions into nine categories using an LLM. For all our experiments, the questions are classified in a zero shot manner by GPT-4o. The categories,\footnote{Notice that the categories are not necessarily mutually exclusive and one question might fit into more than one category.} as shown in Figure \ref{fig:story-specific}, are defined by a human expert after qualitatively examining many evaluation questions. To validate the accuracy of the LLM classification, one human expert manually annotates 150 questions into the nine categories in which GPT-4o scores ~77\% Top-1 match in accuracy. An examination of the misclassified examples reveals that most misclassifications also fall into a closely related category (Figure \ref{fig:confusion-matrix}).

\section{Results: Do LLM judges catch common errors and literary nuance?}
\label{sec:model_human_preference}
In this section, we evaluate the effectiveness of the two-step framework proposed in Section \ref{sec:approach} using human annotation. 

\paragraph{Human Annotation}
We first create a diverse set of translations by machine translating the stories in our datasets with various LLMs and commercial translation service (e.g., Google Translate). The LLMs ranged from small 7B models to GPT-4o and Claude Sonnet 3.5. We also vary different prompting techniques — sentence level, paragraph level, providing extra summary, zero shot, and few shot — to generate diverse translation (see Appendix \ref{app: prompting for diverse translation}).
Human examination confirmed that the generated translation varied widely, ranging from barely fluent Korean with English code-switching behavior of small models to decent translations generated by large proprietary models. 

We then randomly select 20 machine translated paragraph pairs from each of the 10 stories marked gray in Table \ref{tab:dataset} for human annotation (200 paragraphs in total). We also randomly select three evaluation questions corresponding to those paragraphs for annotation for Step 2 (600 questions in total). 

Because of the subjective nature of evaluating literary translation, we create an extensive annotation guideline alongside the rubric so that annotators have the same understanding of the given task. We used the Label Studio platform\footnote{\url{https://labelstud.io/}}  for the annotation. The annotation guideline and annotation interface will be released with the codebase. Table \ref{tab:agreement} shows that the annotators have high inter-annotator agreement. We also ask the annotators to re-annotate a subset of 40 paragraphs after four weeks of the first annotation, and the \textit{intra}-annotator agreement was ~0.7. 

\begin{table}[t]
    \centering
    \small
    \begin{tabular}{l|cccc}
    \toprule
    \textbf{RULER} & \textbf{Kendall's} & \textbf{Spear.} & \textbf{MSE} & \textbf{Krip.} \\
    \textbf{Category} & \textbf{$\tau$} & \textbf{$\rho$} & & \textbf{$\alpha$} \\
    \midrule
    Honorifics & .70 & .74 & .86 & .74 \\
    Lex. Choices & .73 & .81 & .69 & .80 \\
    Syntax & .66 & .76 & .86 & .75 \\
    Content & .73 & .81 & .75 & .80 \\
    \midrule
    VERSE & .65 & .71 & .41 & .70 \\
    \bottomrule
    \end{tabular}
    \caption{\small Inter-annotator agreement metrics across categories. Higher values indicate better agreement for Kendall's $\tau$, Spearman's $\rho$, and Krippendorff's $\alpha$, while lower values are better for MSE. The values are averaged over three annotators. Notice that the last row, VERSE, belonging to Step 2, is measured between 1 to 3 while the top four rows, belonging to Step 1 (RULER), are measured on a 1 to 5 scale.}
    \label{tab:agreement}
\end{table}

\paragraph{Experiment Setting} We compare several open-weight and proprietary models (Table \ref{tab:model-names}) as evaluators for our experiments. We set the temperature to 0.0 for all experiments in this section for reproducibility.

\paragraph{Baseline} According to the common practice of literary machine translation \cite{karpinska2023large, chen2024benchmarkingllmstranslatingclassical, he2024exploring} , we report COMET \cite{rei2022comet}, BERTScore \cite{Zhang2020BERTScore:}, BLEURT \cite{sellam2020bleurt} and GEMBA-MQM \cite{kocmi2023gemba} as the baseline. However, the comparison may not be entirely appropriate because these metrics' designs are quite different from our framework, meant to be used in sentence-level evaluation, while our framework is designed for paragraph-level texts. See Appendix \ref{app: baseline details} for details.

\subsection{RULER: Can LLMs Capture Common Mistakes in Literary Machine Translation?}
\begin{table}[ht]\small
\small
\centering
\setlength{\tabcolsep}{2.5pt}
\begin{tabular}{llcccccc}
\toprule
\textbf{Crit.} & \textbf{Label} & \textbf{Sup.} & \textbf{H-} & \textbf{GPT-4o-} & \textbf{H-} & \textbf{GPT-4o-} \\
& & & \textbf{F1} & \textbf{F1} & \textbf{Acc} & \textbf{Acc} \\
\midrule

\multirow{5}{*}{Honor.} 
& 1 & 30 & 0.79 & 0.07 & \multirow{5}{*}{0.81} & \multirow{5}{*}{0.67} \\
& 2 & 4  & 0.15 & 0.10 & & \\
& 3 & 9  & 0.11 & 0.04 & & \\
& 4 & 9  & 0.22 & 0.14 & & \\
& 5 & 147 & 0.91 & 0.88 & & \\
\midrule
\multirow{5}{*}{Lexical} 
& 1 & 23 & 0.54 & 0.43 & \multirow{5}{*}{0.56} & \multirow{5}{*}{0.48} \\
& 2 & 29 & 0.47 & 0.44 & & \\
& 3 & 42 & 0.45 & 0.37 & & \\
& 4 & 50 & 0.50 & 0.51 & & \\
& 5 & 55 & 0.73 & 0.56 & & \\
\midrule
\multirow{5}{*}{Syntax} 
& 1 & 24 & 0.61 & 0.39 & \multirow{5}{*}{0.52} & \multirow{5}{*}{0.44} \\
& 2 & 32 & 0.55 & 0.44 & & \\
& 3 & 47 & 0.50 & 0.27 & & \\
& 4 & 45 & 0.37 & 0.43 & & \\
& 5 & 50 & 0.62 & 0.58 & & \\
\midrule
\multirow{5}{*}{Content} 
& 1 & 26 & 0.57 & 0.19 & \multirow{5}{*}{0.55} & \multirow{5}{*}{0.47} \\
& 2 & 29 & 0.41 & 0.47 & & \\
& 3 & 42 & 0.45 & 0.22 & & \\
& 4 & 49 & 0.48 & 0.43 & & \\
& 5 & 53 & 0.74 & 0.72 & & \\
\bottomrule
\end{tabular}
\caption{\small Performance comparison across different evaluation criteria. H-F1 represents the average F1-score between human annotator pairs (3 pairs), while LLM-F1 shows the average F1-score between human annotators and GPT-4o (3 pairs). H-Acc and LLM-Acc show the average accuracy for human-human and human-LLM comparisons respectively. Support (Sup.) shows the average number of instances for each label across three annotators (rounded).}
\label{tab:(Step 1) F1 Comparison}
\end{table}
\begin{table*}[t]
\centering
\footnotesize
\setlength{\tabcolsep}{4pt}
\begin{tabular}{l|ccc|ccc|ccc|ccc|ccc}
\toprule
& \multicolumn{3}{c|}{\textbf{Honorifics}} 
& \multicolumn{3}{c|}{\textbf{Lexical}} 
& \multicolumn{3}{c|}{\textbf{Syntax}} 
& \multicolumn{3}{c|}{\textbf{Content}} 
& \multicolumn{3}{c}{\textbf{Step 2}} \\
\cmidrule(lr){2-4} \cmidrule(lr){5-7} \cmidrule(lr){8-10} \cmidrule(lr){11-13} \cmidrule(lr){14-16}
\textbf{Model} & $\tau$ & $\rho$ & MSE 
& $\tau$ & $\rho$ & MSE 
& $\tau$ & $\rho$ & MSE 
& $\tau$ & $\rho$ & MSE 
& $\tau$ & $\rho$ & MSE \\
\midrule
\texttt{gpt-4o} & \textbf{.51} & \textbf{.56} & 2.03 & \textbf{.68} & \textbf{.77} & \textbf{0.72} & \textbf{.62} & \textbf{.71} & \textbf{0.97} & \textbf{.70} & \textbf{.78} & \textbf{0.81} & \textbf{0.49} & \textbf{0.53} & \textbf{0.63} \\
\texttt{claude-3-5-sonnet} & .40 & .41 & 2.22 & .60 & .67 & 1.13 & .59 & .66 & 1.17 & .57 & .64 & 1.35 & 0.38 & 0.41 & \textbf{0.63} \\
\texttt{claude-3-5-haiku} & .42 & .45 & 1.80 & .54 & .59 & 1.35 & .56 & .64 & 1.22 & .55 & .62 & 1.39 & 0.39 & 0.42 & 0.66 \\
\texttt{gpt-4o-mini} & .27 & .31 & 2.74 & .63 & .70 & 0.93 & .59 & .67 & 0.93 & .60 & .67 & 1.08 & 0.37 & 0.40 & \textbf{0.63} \\
\texttt{claude-3-haiku} & .22 & .23 & 2.17 & .45 & .50 & 1.55 & .42 & .47 & 1.47 & .51 & .56 & 1.51 & 0.19 & 0.20 & 0.87 \\
\texttt{gemma-2-27b-it} & .34 & .38 & 2.08 & .47 & .52 & 1.27 & .49 & .56 & 1.24 & .49 & .55 & 1.25 & 0.18 & 0.19 & 0.70 \\
\texttt{gemma-2-9b-it} & .28 & .30 & 2.36 & .43 & .48 & 1.40 & .36 & .41 & 1.47 & .52 & .58 & 1.34 & 0.09 & 0.09 & 0.73 \\
\texttt{llama-3-8b-it} & -.01 & -.01 & 2.31 & .37 & .41 & 1.56 & .29 & .32 & 1.65 & .21 & .24 & 2.04 & 0.05 & 0.06 & 1.06 \\
\midrule
COMET & .10 & .12 & -- & .28 & .35 & -- & .25 & .32 & -- & .28 & .36 & -- & 0.24 & 0.31 & -- \\
BERTScore & .12 & .15 & -- & .23 & .29 & -- & .22 & .28 & -- & .24 & .30 & -- & 0.20 & 0.27 & -- \\
BLEURT & .14 & .17 & -- & .23 & .30 & -- & .21 & .28 & -- & .23 & .30 & -- & 0.20 & 0.26 & -- \\
GEMBA-MQM (gpt-4o) & .26 & .32 & -- & .49 & .63 & -- & .48 & .61 & -- & .50 & .63 & -- & 0.38 & 0.46 & -- \\
\bottomrule
\end{tabular}
\caption{\small{Evaluation results on RULER (Step 1) and VERSE (Step 2). Step 1 reports Kendall’s $\tau$, Spearman’s $\rho$, and mean squared error (MSE) across four fine-grained criteria: honorifics, lexical choice, syntax, and content. Step 2 provides overall correlation with human judgments of whether each criterion was satisfied. Higher is better for $\tau$ and $\rho$, and lower is better for MSE; Overall, larger models consistently outperform smaller ones, though performance varies across criteria and evaluation settings. Table \ref{tab:step1_2_results_cot} includes some newer models.}}
\label{tab:step1_2 main table}
\end{table*}

We first prompt the language model with the human-curated rubrics for each criterion and ask it to give a score. The LLM is also given one gold translation for reference.  
As shown in Table \ref{tab:step1_2 main table}, across all the criteria, our framework outperforms traditional metrics and GPT-4o based GEMBA-MQM, however, even the best evaluator, GPT-4o does not match human-level correlation reported in Table \ref{tab:agreement}.

\paragraph{Honorific Failure Mode} A careful analysis reveals that GPT-4o and all other models \textit{completely} fail to notice serious mistakes in honorifics. Table \ref{tab:(Step 1) F1 Comparison} shows that GPT-4o often gives high scores for translations that received low scores from humans regarding honorifics (notice the low F1 score for labels 1, 2 and 3).

Korean honorifics play a crucial role in dialogue-based literary texts, drastically impacting the reading experience when improperly translated. Our evaluation rubric assigns a default score of 5 to non-dialogue passages, which means a significant number of passages automatically get 5. Thus, in our case, it is important for the models to detect when the honorifics get translated \textit{incorrectly}. However, as Figure \ref{fig:Step 1 Honorifics analysis} shows GPT-4o's predicted label distribution does not match well with human-to-human agreement. In contrast, GPT-4o does a better job at matching human preference distribution for Syntax, Content Accuracy, and Lexical Choices. However, it still tends to provide overoptimistic estimations, captured by higher MSE compared to inter-human. We explore this phenomenon further in Section \ref{sec: framework-use-cases}. 


\subsection{VERSE: Can LLMs Answer Literary Questions with Subtle Nuance?}
Similar to Step 1 RULER, we prompt the language models with human curated rubric and a reference human translation and ask them to give a score.

As shown in Table \ref{tab:step1_2 main table}, language models are much better at correlating with human preference in understanding literary translation than the traditional MT metrics. However, there remains a gap between expert human-level agreement (Table \ref{tab:agreement}). In-depth analysis of GPT-4o (the best-performing model above) performance shows that the \textbf{model has a particularly lower recall than inter-annotators for Score 1 than for Score 2 and 3} (Table \ref{tab:(Step 2) classification-comparison}), which means GPT-4o cannot effectively spot translations that critically fail to satisfy specific criteria. However, the recall drop was not as significant as in the case of honorifics in Step 1 Honorific criteria.

\begin{table}[t]
\small
\centering
\setlength{\tabcolsep}{3pt}
\begin{tabular}{cc|ccc|ccc}
\toprule
& & \multicolumn{3}{c|}{\textbf{Human-Human}} & \multicolumn{3}{c}{\textbf{Human-Model}} \\
\cmidrule(lr){3-5} \cmidrule(lr){6-8}
\textbf{Label} & \textbf{Support} & \textbf{Prec} & \textbf{Rec} & \textbf{F1} & \textbf{Prec} & \textbf{Rec} & \textbf{F1} \\
\midrule
1 & 197/189 & 0.81 & 0.75 & 0.78 & 0.77 & 0.47 & 0.59 \\
2 & 182/171 & 0.52 & 0.46 & 0.49 & 0.38 & 0.51 & 0.43 \\
3 & 185/203 & 0.64 & 0.78 & 0.70 & 0.59 & 0.65 & 0.62 \\
\midrule
Avg & 188/188 & 0.66 & 0.66 & 0.66 & 0.58 & 0.54 & 0.55 \\
\bottomrule
\end{tabular}
\caption{\small{Comparison of classification metrics between human annotators (Human-Human) and GPT-4o no CoT predictions (Human-Model). Support shows the average number of instances for human/model evaluations.}}
\label{tab:(Step 2) classification-comparison}
\end{table}
\subsection{Ablation of Different Components in the Framework}

A set of ablations tables in Appendix \ref{app:additional-results} shows that:
\begin{itemize}
    \item \textbf{Chain-of-thought (CoT) does not help:} Adding CoT reasoning generally fails to improve evaluation quality in either RULER or VERSE; in some cases it even slightly reduces performance. This holds true for even the `thinking' reasoning models. See Table \ref{tab:step1_2_results_cot}. 
    \item \textbf{A clear grading rubric matters more than references:} Removing the rubric hurts results much more than removing the human reference translation, showing that structured evaluation guidelines are essential (Table \ref{tab:(step 1_2) rubric and human tr ablation}).
    \item \textbf{Different effects across steps:} In Step 1 (RULER), human references slightly help, specially with honorifics; but in Step 2 (VERSE), models often perform better without references, suggesting different dynamics in rubric-based error detection vs. nuanced question answering (Table \ref{tab:(step 1_2) rubric and human tr ablation}).
    \item \textbf{Not much gain from few-shot prompting:} Providing few-shot examples does not consistently boost performance in either step, and in some cases decreases it (Table \ref{tab:Step 1_2 kshot-performance}).
\end{itemize}
\section{Discussion: Promise and Pitfall of the Proposed Framework}
\label{sec: framework-use-cases}
While larger language models perform clearly better than traditional machine translation metrics, they still fall behind inter-human correlation. We, therefore, further examine what the use cases and limitations of our framework for automated literary translation evaluation may be. 

\subsection{Experiment Setting}
For all the experiments in this section, we use the GPT-4o as the evaluator model for all experiments in zero-shot no CoT setting.\footnote{For VERSE, Claude-Sonnet 3.5 is slightly better using CoT, however, using chain of thought becomes prohibitively expensive for large scale evaluation, which is why we resort to zero shot no CoT GPT-4o, whose performance is very close.} We also did not use reference human translation because in Step 2 reference-free configuration performs slightly better.

\paragraph{Evaluated Translations} To test the efficacy of our framework, we use several language models and Google Translate to translate 10 stories from our dataset. Following the best practices of \cite{karpinska2023large}, we prompt language models with five examples and then generate translations. We also append a GPT-4o generated summary of the story for context. 


\subsection{VERSE Provides Fine Grained Literary Evaluation}

\begin{figure}[t]
    \includegraphics[width=0.45\textwidth]{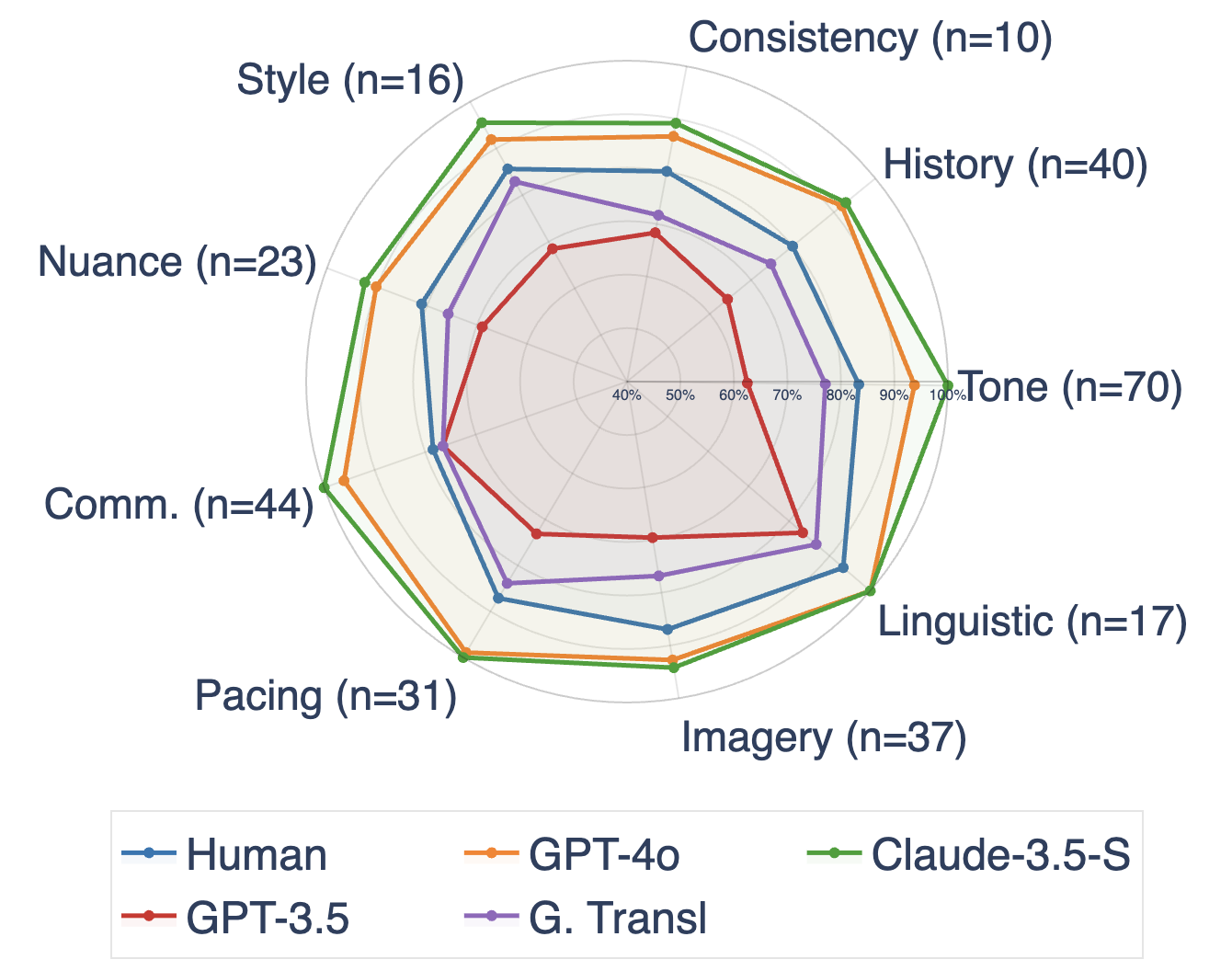}
    \caption{\small{Radar plot for scores (as percentage) across different categories of translation for each model in The Philosopher story. The radial axis begins at 40\%. See Figure \ref{fig:step 2 all stories evaluated} for more examples.}}
    \label{fig:Radar Plot The Philosopher}
\end{figure}

Step 2 (VERSE) of our framework permits a novel fine-grained interpretation into literary translation, which is otherwise hard to capture through a single number or traditional machine translation metric like fluency, adequacy, etc. Figure \ref{fig:Radar Plot The Philosopher} shows one such use case of such approach to quickly inspect and visualize different dimensions present in the translation. Moreover, because the language model evaluates ~10 questions per paragraph, VERSE estimations are also robust.

\subsection{LLMs prefers LLM generated translation}
Table \ref{tab: (Step 1) averaged over all stories} shows the aggregated scores for each Step 1 criteria for each model. Note that since our proposed framework can be run in reference free fashion, we evaluate the reference translations as well. 

The results are surprising. Overall, Claude-3.5 Sonnet (as translator) gets the highest scores, followed by GPT-4o, with almost perfect scores in several dimensions. Human translation receives \textit{lower} scores than the bigger models. To better understand the cause, we sample 50 paragraphs from the story ``What is Remembered'', whose human translation receive a particularly lower score than the best performing LLM translation by Claude-Sonnet 3.5. In a controlled A/B setting, one human annotator picked the gold translation as the better one 39 out of 50 times, and the Claude-Sonnet 3.5 translation was picked only 4 times.

One potential reason why evaluator LM is rating human translation lower might have to do with the nature of human translations. Human translations often deviate significantly from their original text in order to properly convey the linguistic nuances. Human inspection reveals that this is indeed the case for the ``What is Remembered" story. Moreover, for all of our stories, a casual inspection reveals that human translations are clearly better than the best machine translations in majority of the cases.

We observe the same for Step 2, where translations by large models are rated higher than human translations.

This suggests that \textbf{while current LLMs might be good at spotting significant differences in translation qualities, they still often fail to distinguish translations where the difference is more subtle and nuanced}. This can be observed from the relatively lower scores provided to Google Translate and GPT-3.5, which usually produce very poor Korean translation as seen upon human inspection.

\begin{table}[t]
\centering
\footnotesize
\setlength{\tabcolsep}{4pt}
\begin{tabular}{l|cccc|c}
\toprule
\textbf{Transl. Model } & \textbf{Hon.} & \textbf{Syn.} & \textbf{Lex.} & \textbf{Con.} & \textbf{Step 2} \\
\midrule
\texttt{Human} & 98.97 & 88.21 & 89.74 & 83.08 & 89.62 \\
\texttt{GPT-4o} & 99.49 & 97.44 & 94.36 & \textbf{97.95} & 92.41 \\
\texttt{Claude-3.5-Son.} & \textbf{100.00} & \textbf{98.97} & \textbf{98.97} & \textbf{98.97} & 95.55 \\
\texttt{Claude-3.5-Hai.} & \textbf{100.00} & 94.87 & 96.41 & 96.92 & 92.76 \\
\texttt{Claude-3-Haiku} & 98.46 & 88.72 & 85.64 & 89.23 & 88.11 \\
\texttt{GPT-4-Mini} & 99.49 & 90.26 & 88.21 & 92.31 & 87.51 \\
\texttt{GPT-3.5} & 91.79 & 57.44 & 56.92 & 60.00 & 67.56 \\
\texttt{Google Trans.} & 97.95 & 71.79 & 69.23 & 76.41 & 75.53 \\
\bottomrule
\end{tabular}
\caption{\small Average scores (in percentage) across different models on four criteria of translation quality, computed over the 10 evaluation stories (501 paragraphs). Note that while LLMs can discriminate between very good and very bad translations (eg, GPT-4o vs GPT-3.5), its ability to distinguish between high-quality translations is limited. The rightmost column is averaged scores on Step 2 VERSE evaluation.}
\label{tab: (Step 1) averaged over all stories}
\end{table}


\paragraph{LLM's tendency to overestimate scores}
\begin{table*}[ht]
\centering
\scriptsize 
\setlength{\tabcolsep}{3pt} 
\begin{tabular}{l|cccc|c|ccccccccc|c}
\toprule
\textbf{Translation Model} & \textbf{Hon.} & \textbf{Syn.} & \textbf{Lex.} & \textbf{Con.} & & \textbf{Style} & \textbf{Char.} & \textbf{Hist.} & \textbf{Img.} & \textbf{Comm.} & \textbf{Ling.} & \textbf{Pace} & \textbf{Nuance} & \textbf{Cons.} & \textbf{Mean} \\
\midrule
\texttt{gpt-4o} & 99.49 & 97.44 & 94.36 & 97.95 & \vline & 89.70 & 92.16 & 90.58 & 90.76 & 93.09 & 88.36 & 96.42 & 92.00 & 98.64 & 92.41 \\
\rowcolor{gray!15} ~~w/ Ref & 98.97 & 86.67 & 82.56 & 90.77 & \vline & 93.21 & 94.68 & 90.12 & 92.28 & 94.87 & 90.42 & 98.02 & 94.04 & 99.42 & 94.12 \\
\midrule
\texttt{claude-3.5-sonnet} & 100.00 & 98.97 & 98.97 & 98.97 & \vline & 94.42 & 95.35 & 92.51 & 94.12 & 96.12 & 93.21 & 98.91 & 95.29 & 100.00 & 95.55 \\
\rowcolor{gray!15} ~~w/ Ref & 99.49 & 95.90 & 90.77 & 96.92 & \vline & 95.88 & 96.87 & 92.28 & 95.07 & 96.64 & 94.42 & 99.43 & 96.71 & 100.00 & 96.37 \\
\midrule
\texttt{claude-3.5-haiku} & 100.00 & 94.87 & 96.41 & 96.92 & \vline & 90.30 & 92.46 & 90.47 & 91.70 & 93.53 & 87.39 & 97.13 & 92.86 & 99.03 & 92.76 \\
\rowcolor{gray!15} ~~w/ Ref & 98.97 & 87.69 & 82.56 & 91.28 & \vline & 94.18 & 94.93 & 90.00 & 93.36 & 94.87 & 89.58 & 98.47 & 94.43 & 99.61 & 94.38 \\
\midrule
\texttt{claude-3-haiku} & 98.46 & 88.72 & 85.64 & 89.23 & \vline & 85.82 & 87.06 & 86.84 & 85.33 & 89.40 & 80.61 & 93.93 & 87.45 & 96.51 & 88.11 \\
\rowcolor{gray!15} ~~w/ Ref & 96.92 & 80.51 & 76.92 & 81.54 & \vline & 88.48 & 89.67 & 85.38 & 86.68 & 90.55 & 83.52 & 94.83 & 89.10 & 97.29 & 89.50 \\
\midrule
\texttt{gpt-4-mini} & 99.49 & 90.26 & 88.21 & 92.31 & \vline & 82.79 & 87.15 & 85.91 & 85.73 & 88.06 & 81.82 & 92.02 & 87.37 & 96.71 & 87.51 \\
\rowcolor{gray!15} ~~w/ Ref & 98.97 & 80.51 & 75.90 & 86.15 & \vline & 87.15 & 90.09 & 85.44 & 87.21 & 90.70 & 84.12 & 94.70 & 89.73 & 98.64 & 89.75 \\
\midrule
\texttt{gpt-3.5} & 91.79 & 57.44 & 56.92 & 60.00 & \vline & 64.48 & 67.03 & 67.54 & 65.68 & 68.06 & 63.27 & 69.28 & 67.29 & 75.39 & 67.56 \\
\rowcolor{gray!15} ~~w/ Ref & 88.21 & 51.79 & 51.28 & 54.87 & \vline & 64.61 & 66.70 & 64.91 & 63.53 & 67.91 & 62.55 & 68.97 & 67.22 & 72.87 & 66.58 \\
\midrule
\texttt{google-translate} & 97.95 & 71.79 & 69.23 & 76.41 & \vline & 71.39 & 73.14 & 74.68 & 74.79 & 76.21 & 68.00 & 79.18 & 75.76 & 86.63 & 75.53 \\
\rowcolor{gray!15} ~~w/ Ref & 95.90 & 66.15 & 63.59 & 68.72 & \vline & 72.24 & 74.42 & 73.04 & 74.07 & 75.78 & 67.64 & 78.86 & 74.90 & 84.88 & 75.09 \\
\midrule\midrule
\texttt{human} & 98.97 & 88.21 & 89.74 & 83.08 & \vline & 88.48 & 89.40 & 87.02 & 88.29 & 90.84 & 84.12 & 94.06 & 88.78 & 95.54 & 89.62 \\
\bottomrule
\end{tabular}
\caption{\small Aggregated scores for the ten stories on RULER (Hon. - Con.) and VERSE (Style - Mean). Mean only refers to the mean of the second block, VERSE.  Gray rows indicate aggregated scores when the reference human translation was also provided during evaluation. Scores for human translation were calculated in reference free manner. The model name represents the model that generated the translation. Inclusion of reference translations leads to lower scores in RULER, whereas it yields higher scores in VERSE. More interestingly, human translations are rated poorer than some machine translations. }
\label{tab:(Step 1_2) avg over all stories with reference}
\end{table*}
Table \ref{tab:(Step 1_2) avg over all stories with reference} shows that evaluator LLM GPT-4o tends to overestimate the quality of LLM generated translations over human translations. An interesting pattern is observed with the \textit{direction} of overestimation. For RULER (Step 1) scores, providing reference translation decreases the average score, while for LLM generated QA in VERSE, the average score increases. This is in line with the results observed in the ablations in Table \ref{tab:(step 1_2) rubric and human tr ablation} where MSE increases when reference is not provided for RULER (Step 1), and decreases for VERSE (Step 2).

Notice, however, that the relative ranks of the models do change in a few cases in these two settings, especially in the middle ranks. The best and worst-performing models' ranks do not change. 

This observation supports our previous observation in Section \ref{sec: framework-use-cases} that GPT-4o as a judge can differentiate between obviously good and obviously bad translations, however, its capability to detect differences in similarly (high) quality machine translations is not yet as robust. 

\paragraph{Generalizability of Our Framework} 
Although our study focuses on English–Korean, the two-step framework is adaptable to other language pairs. To provide interpretability suited specifically for literary translation, the framework introduces a novel design that combines a general, language-agnostic component with a culturally-specific rubric-based evaluation. VERSE requires little to no modification and can operate with minimal human supervision. RULER, though it operates on language-specific rubrics, can be adapted to other languages with adjustments that reflect their particular linguistic and cultural contexts.



\section{Conclusion}
In this work, we propose and evaluate the feasibility of using LLM as an automated tool for fine-grained evaluation of literary text and found that current LLMs, while getting closer to human-level judgment than traditional metrics and providing useful interpretable insights, still lag behind human judgments, especially in dimensions like honorifics. The findings call for more research in developing metrics for evaluating long-form translations, especially in the literary domain.

\section*{Limitations and Future Work}
This study evaluates the framework only on English–Korean. A central challenge for future work is to test how well LLMs can generate literary-specific questions in non-English settings and answer them in ways that respect the cultural conventions of the target language. It is possible that LLMs will not be able to address high-quality questions in low-resource languages or languages that they haven’t been trained on extensively. Similarly, it is also possible that the framework works even better in higher-resource languages. Applying our two-step framework to diverse language pairs will further reveal its capacity to capture the rich cultural variation that defines literary translation.

\bibliography{main}
\bibliographystyle{acl_natbib}
\newpage
\appendix
\label{sec:appendix}






\section{Dataset Detail}
\label{app: dataset}

The dataset detail can be found in Table \ref{tab:dataset}.

\begin{table*}[t]
\centering
\footnotesize
\begin{tabular}{@{}p{0.20\textwidth}p{0.15\textwidth}cp{0.25\textwidth}cc@{}}
\toprule
\textbf{Title} & \textbf{Author} & \textbf{Para. Count} & \textbf{Translator(s)} & \textbf{Orig.} & \textbf{Trans.} \\
\midrule
\rowcolor{gray!20}
Austin & Andrew J. Porter & 26 & Eun Young Min & 2017 & 2024 \\
\rowcolor{gray!20}
The Passing of Grandison & Charles W. Chesnutt & 56 & Ki Wook Han & 1899 & 2010 \\
\rowcolor{gray!20}
The Ones Who Walk Away\newline from Omelas & Ursula K. Le Guin & 14 & Yong Jun Choi & 1973 & 2014 \\
\rowcolor{gray!20}
The Garden Party & Katherine Mansfield & 44 & Young Hee Kim & 1922 & 2010 \\
\rowcolor{gray!20}
Real Artists & Ken Liu & 84 & Seong Ju Jang & 2011 & 2024 \\
\rowcolor{gray!10}
The Philosopher & Sherwood Anderson & 29 & Seon H. Kim, Young W. Park & 1919 & 16, 19 \\
\rowcolor{gray!10}
A Retrieved Reformation & O. Henry & 36 & W. D. Kim, Jeong A Ko & 1903 & 03, 14 \\
\rowcolor{gray!10}
Amnesty & Octavia Butler & 84 & Soo Hyun Lee & 2003 & 2016 \\
\rowcolor{gray!10}
Tower of Babylon & Ted Chiang & 107 & Sang Hoon Kim & 1990 & 2016 \\
\rowcolor{gray!10}
The Red Masque of Death & Edgar Allan Poe & 14 & D. H. Jeon, S. H. Jeon & 1842 & 09, 14 \\
Hell-Heaven & Jhumpa Lahiri & 34 & Sang Mi Park & 2004 & 2009 \\
The Legacy & Virginia Woolf & 16 & Young Hee Kim & 1944 & 2010 \\
The Signal Man & Charles Dickens & 52 & Young Hee Kim & 1866 & 2010 \\
What is Remembered & Alice Munro & 90 & Jeong Eun Seo & 2001 & 2020 \\
The Deposition & Tobias Wolff & 39 & Jae Kyung Lee & 2006 & 2012 \\
\bottomrule
\end{tabular}
\caption{\small The details of the stories being used in our work. The number of paragraphs and when they were published are shown. Three stories in our corpus have two different translations, for which we included the names of both translators. Gray row stories were used for human evaluation experiments in Section \ref{sec:model_human_preference}. \colorbox{gray!20}{Dark gray} on top were used to create the few-shot examples for all the experiments in Section \ref{sec:approach} and \ref{sec: framework-use-cases}. The other five \colorbox{gray!10}{lighter gray} works were part of the human evaluation in \ref{sec:approach} but their paragraphs were not part of the few-shot examples. We leave the five other stories solely for the purpose of our analysis in Section \ref{sec: framework-use-cases}.}
\label{tab:dataset}
\end{table*} 
\section{Model Name Details}
\label{app: model names}


\begin{table}[t]
\centering
\scriptsize
\setlength{\tabcolsep}{3pt}     
\renewcommand{\arraystretch}{1.05}
\begin{tabularx}{\columnwidth}{L R}
\toprule
\textbf{Model} & \textbf{Exact version string} \\
\midrule
GPT-4o \citep{openai_gpt4o_2024}               & \texttt{gpt-4o-2024-05-13} \\
GPT-4.1 \citep{openai_gpt41_2025}              & \texttt{gpt-4.1-2025-04-14} \\
o3 \citep{openai_o3_2025}                      & \texttt{o3-2025-04-16} \\
Claude-3.5 Sonnet \citep{anthropic_claude35_sonnet_2024}
                                               & \texttt{claude-3-5-sonnet-20240620} \\
Claude-3.5 Haiku \citep{anthropic_claude35_haiku_2024_addendum}
                                               & \texttt{claude-3-5-haiku-20241022} \\
GPT-4o Mini \citep{openai_gpt4omini_2024}      & \texttt{gpt-4o-mini-2024-07-18} \\
Claude-3 Haiku \citep{anthropic_claude3_family_2024}
                                               & \texttt{claude-3-haiku-20240307} \\
Gemma 2 (27B) \citep{gemma2_2024}              & \texttt{gemma-2-27b-it} \\
Gemma 2 (9B) \citep{gemma2_2024}               & \texttt{gemma-2-9b-it} \\
Llama 3.1 (8B) \citep{llama3_herd_2024,meta_llama31_2024_blog}
                                               & \texttt{Llama-3.1-8B-Instruct} \\
DeepSeek V3 \citep{deepseek_v3_2024}           & \texttt{deepseek-chat-v3} \\
DeepSeek R1 \citep{deepseek_r1_2025}           & \texttt{deepseek-r1} \\
\bottomrule
\end{tabularx}
\caption{LLMs and exact versions used in the evaluation experiments.}
\label{tab:model-names}
\end{table}

Table \ref{tab:model-names} shows the names and the details of the models used in our work.

\subsection{Baseline Details}
\label{app: baseline details}
For the BERTScore, BLEURT, COMET baselines, we use the following versions used in the previous work \cite{karpinska2023large}:
\paragraph{BERTScore: } We used a \texttt{roberta-large} pretrained checkpoint from huggingface. 
\paragraph{BLEURT: } We used \texttt{bleurt-20} checkpoints from the official repository.
\paragraph{COMET:} \texttt{wmt22-comet-da} checkpoints.
\paragraph{GEMBA-MQM:} We used the reference free mode of GEMBA-MQM \cite{kocmi2023gemba} since the original paper claims it to be superior to the mode using reference translation. We used the same GPT4o model that powers most of our experiments.

Note that GEMBA-MQM and our framework's numbers are not really comparable since the MQM categories and our categories are very different.

\section{Prompting Methods Used for Generating Diverse Translations}
\label{app: prompting for diverse translation}

To obtain diverse translations for the human evaluation stage (Section \ref{sec:model_human_preference}), we explored several prompting setups with both large and small LLMs. Translations were generated either sentence by sentence or at the paragraph level, and we varied between zero-shot and few-shot settings. Consistent with prior work \cite{zhang-etal-2023-machine}, as well as our own inspection, few-shot prompting generally produced higher-quality outputs. In some cases we also prepended an LLM-generated summary to the few-shot examples. For the final analysis (§\ref{sec: framework-use-cases}), we used five-shot prompting with a GPT-4o summary. The examples were drawn from stories other than the one being evaluated. We made sure to include both dialogue- and narrative-based paragraphs. All translations were generated with the models’ default temperature (unlike evaluation, which used $T=0.0$).

\subsection{LLM Generated Questions to Evaluate Literary Translations}
\subsubsection{Categories} 

These are the original human-curated categories of the LLM-generated questions:

"Historical and Cultural Context", "Imagery and Descriptive Quality", "Character Voice, Tone, and Individuality", "Interpersonal Communication and Hierarchy", "Linguistic and Idiomatic Naturalness", "Nuanced Interpretation including Subtle Implications", "Narrative Pacing and Rhythm", "Affective and Stylistic Resonance", "Overall Consistency and Cohesion"

The categories have been abridged on all the plots for the sake of visualization.  



\section{Details of Common Issues found in Machine Translation}
\label{appendix:common_issues}
\paragraph{Word/Phrase/Expression Level Issues} This type of error occurs most frequently and significantly impacts overall translation quality. While LLMs excel at literal translations, they struggle with natural word choices in the target language, particularly when dealing with figurative expressions. This often results in the selection of awkward or uncommon verbs in Korean; although the overall content remains comprehensible, these choices critically diminish the quality of translation. Notably, problems often arise with polysemous words or idiomatic expressions, seriously impeding reader comprehension. Additional common errors include inappropriate use of pronouns, such as referring to people as “\korean{그것} (it),” reflecting the essential difference in the linguistic structure between English and Korean.

\paragraph{Context Related Issues} This type of error presents more complex and serious challenges arising from cultural and historical differences between the source and target languages. For instance, because terms like “abolitionist” in “The Passing of Grandison” do not have an equivalent historical term in Korean language, LLMs end up offering incomprehensible expressions. Similarly, the expression “border states” often loses its geographical context (mistranslated as “\korean{국경주}–states close to national borders” in Korean). LLMs also struggle with specialized terminology and historical references, as shown in mistranslations of terms like “bootjack” and “nightcap.”

\paragraph{Form-related issues} This third category includes subcategories concerning punctuation and grammar, as well as structure and formatting. Punctuation and grammar refer to problems with punctuation changes that alter meaning. For example, when LLM translation replaces semicolons with periods, it alters the original text's meaning. Likewise, changes in the sentence structure in the translated text can lead to incomplete or awkward results. This subcategory includes problems with splitting long sentences unnecessarily, which results in awkward structures when handling complex sentences. Other issues in this category included structure and formatting. These appeared when formatting the dialogue and narrative flows.  The incorrect handling of sentence structures in the LLM translation led to awkward sentences, negatively affecting readability. Form-related issues also include transliteration, consistency issues, and omissions.  LLMs transliterate Western names inconsistently in the translation, resulting in different name variations throughout the translated text. (i.e., translating “Grandison” as “\korean{그라디슨}" and “\korean{그란디손}” within a single translation) LLM translations may eliminate a sentence and expression or add arbitrary content different from the original text. They also leave certain words untranslated, sometimes including various words or phrases in random languages within the translation. 

\paragraph{Honorifics} Errors in this category are notable in English-to-Korean LLM translations regarding features of Korean, in which different levels of formality are presented through various verb endings. Mistranslation of honorifics and formality appear most frequently in dialogues, reflecting a misinterpretation of the relationships between characters, or inconsistent use of honorifics within a single character’s speech. Translation issues in the honorifics category critically affect the reading experience because it encompasses naturalness, as well as proper understanding of character power dynamics and subtle psychological contexts.


\newpage
\section{Additional Results and Ablations}
\label{app:additional-results}

\begin{figure}[ht]
    \centering
    \begin{subfigure}[b]{0.48\columnwidth}
        \centering
        \includegraphics[width=\linewidth]{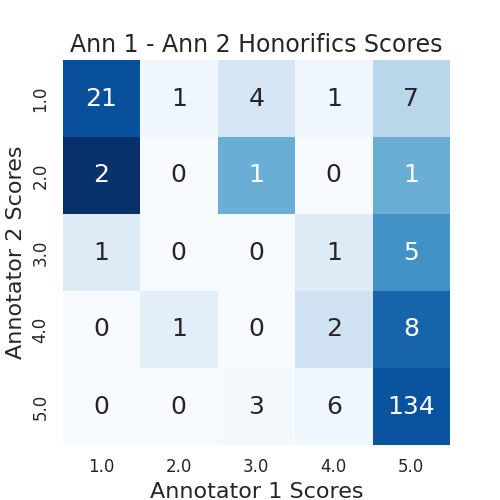}
        \caption{}
        \label{fig:ann1-2-honorifics}
    \end{subfigure}
    \hfill
    \begin{subfigure}[b]{0.48\columnwidth}
        \centering
        \includegraphics[width=\linewidth]{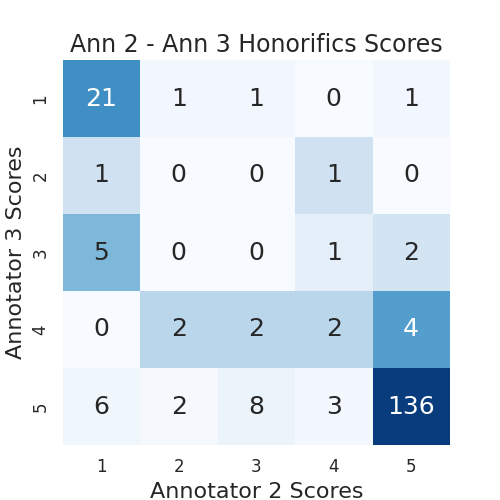}
        \caption{}
        \label{fig:ann2-3-honorifics}
    \end{subfigure}
    
    \begin{subfigure}[b]{0.48\columnwidth}
        \centering
        \includegraphics[width=\linewidth]{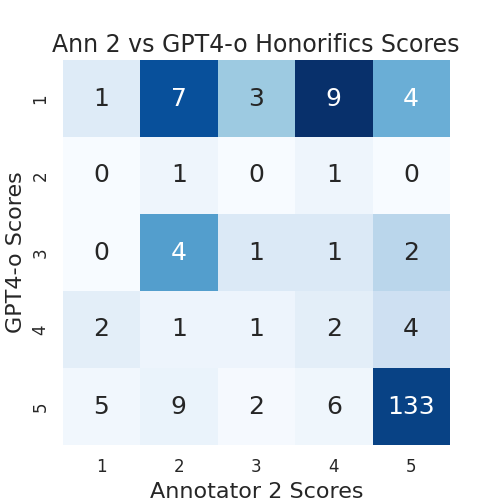}
        \caption{}
        \label{fig:ann2-gpt4-honorifics}
    \end{subfigure}
    \hfill
    \begin{subfigure}[b]{0.48\columnwidth}
        \centering
        \includegraphics[width=\linewidth]{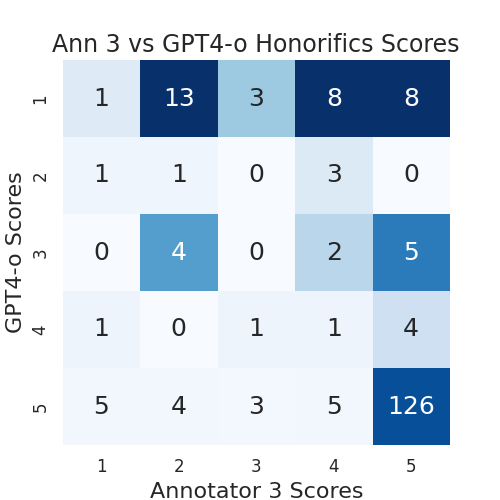}
        \caption{}
        \label{fig:ann3-gpt4-honorifics}
    \end{subfigure}
    \caption{\small Confusion matrices for Korean honorifics show strong agreement between human annotators (top row), who consistently mark some paragraphs as very poor (dark blue upper-left corner). In contrast, GPT-4o fails to capture this pattern (lighter upper-left corner, bottom row).}
    \label{fig:Step 1 Honorifics analysis}
\end{figure}

\begin{sidewaystable}[h]
\centering
\scriptsize
\setlength{\tabcolsep}{4pt}
\begin{tabular}{l|c|ccc|ccc|ccc|ccc|ccc}
\toprule
& & \multicolumn{3}{c|}{\textbf{Honorifics}} & \multicolumn{3}{c|}{\textbf{Lexical}} & \multicolumn{3}{c|}{\textbf{Syntax}} & \multicolumn{3}{c|}{\textbf{Content}} & \multicolumn{3}{c}{\textbf{Step 2 VERSE}} \\
\cmidrule(lr){3-5} \cmidrule(lr){6-8} \cmidrule(lr){9-11} \cmidrule(lr){12-14} \cmidrule(lr){15-17}
\textbf{Model} & $k$ & $\tau$ & $\rho$ & MSE & $\tau$ & $\rho$ & MSE & $\tau$ & $\rho$ & MSE & $\tau$ & $\rho$ & MSE & $\tau$ & $\rho$ & MSE \\
\midrule
\multirow{5}{*}{\texttt{gpt-4o}} 
& 0  & \textbf{.51} & \textbf{.56} & \textbf{2.03} & .68 & .77 & \textbf{0.72} & .62 & .71 & \textbf{0.97} & \textbf{.70} & \textbf{.78} & \textbf{0.81} & 0.49 & 0.53 & 0.63 \\
& 5  & .49 & .53 & 1.91 & .68 & .77 & 0.76 & .62 & .71 & 0.98 & .68 & .76 & 0.79 & 0.47 & 0.52 & 0.64 \\
& 10 & .50 & .53 & 1.93 & \textbf{.70} & \textbf{.78} & 0.73 & \textbf{.64} & \textbf{.73} & 0.99 & .65 & .73 & 0.88 & \textbf{0.51} & \textbf{0.55} & \textbf{0.59} \\
& 15 & .49 & .53 & 2.07 & .69 & .78 & 0.73 & \textbf{.64} & \textbf{.73} & 1.03 & .66 & .75 & 0.86 & 0.48 & 0.53 & 0.61 \\
& 20 & .44 & .48 & 2.39 & .67 & .77 & 0.80 & \textbf{.64} & \textbf{.73} & 0.97 & .66 & .75 & 0.89 & 0.49 & 0.53 & 0.63 \\
\midrule
\multirow{5}{*}{\texttt{gpt-4o-mini}} 
& 0  & .27 & .31 & \textbf{2.74} & \textbf{.63} & \textbf{.70} & \textbf{0.93} & .59 & .67 & \textbf{0.93} & .60 & .67 & 1.08 & \textbf{0.37} & \textbf{0.40} & \textbf{0.63} \\
& 5  & .31 & .35 & 3.28 & .55 & .62 & 1.19 & .60 & .67 & \textbf{0.93} & .60 & .67 & 1.00 & 0.21 & 0.23 & 0.75 \\
& 10 & .31 & .34 & 3.20 & .54 & .61 & 1.30 & .54 & .61 & 1.10 & .60 & .67 & \textbf{0.99} & 0.15 & 0.16 & 0.75 \\
& 15 & \textbf{.34} & \textbf{.38} & 3.03 & .49 & .56 & 1.36 & .49 & .56 & 1.22 & \textbf{.64} & \textbf{.71} & \textbf{0.90} & 0.23 & 0.25 & 0.72 \\
& 20 & .26 & .29 & 3.05 & .53 & .59 & 1.35 & .50 & .57 & 1.19 & .59 & .66 & 1.01 & 0.20 & 0.22 & 0.78 \\
\midrule
\multirow{5}{*}{\texttt{llama-3-8b-it}} 
& 0  & -.01 & -.01 & 2.31 & \textbf{.37} & \textbf{.41} & 1.56 & \textbf{.29} & \textbf{.32} & \textbf{1.65} & .21 & .24 & 2.04 & 0.05 & 0.06 & 1.06 \\
& 5  & .19 & .20 & 2.27 & .21 & .23 & 1.93 & .12 & .14 & 2.16 & \textbf{.24} & \textbf{.26} & \textbf{1.98} & 0.05 & 0.05 & 1.53 \\
& 10 & .11 & .12 & 2.19 & .13 & .14 & 2.01 & .14 & .16 & 2.17 & .14 & .16 & 2.12 & \textbf{0.09} & \textbf{0.10} & 1.27 \\
& 15 & .03 & .03 & 2.27 & .18 & .20 & 1.97 & .16 & .18 & 2.14 & .14 & .16 & 2.14 & 0.06 & 0.07 & 0.82 \\
& 20 & \textbf{.25} & \textbf{.26} & \textbf{2.18} & .25 & .28 & \textbf{1.85} & .20 & .22 & 2.03 & .24 & .26 & 2.00 & 0.06 & 0.06 & \textbf{0.77} \\
\bottomrule
\end{tabular}
\caption{\small (Performance comparison when k shot examples were provided (in no COT setting) in RULER metrics and VERSE (Step 2). The best results for each model are shown in \textbf{bold}. In most cases providing few shot examples do not consistently boost model performance and for a few cases it even seems to decrease the performance.}
\label{tab:Step 1_2 kshot-performance}
\end{sidewaystable}
\begin{table*}[h]
\centering
\captionsetup{font=small}
\footnotesize
\setlength{\tabcolsep}{4pt}
\begin{tabular}{l|ccc|ccc|ccc|ccc|ccc}
\toprule
& \multicolumn{3}{c|}{\textbf{Honorifics}} & \multicolumn{3}{c|}{\textbf{Lexical}} & \multicolumn{3}{c|}{\textbf{Syntax}} & \multicolumn{3}{c|}{\textbf{Content}} & \multicolumn{3}{c}{\textbf{Step 2 (VERSE)}} \\
\cmidrule(lr){2-4} \cmidrule(lr){5-7} \cmidrule(lr){8-10} \cmidrule(lr){11-13} \cmidrule(lr){14-16}
\textbf{Eval Model} & $\tau$ & $\rho$ & MSE & $\tau$ & $\rho$ & MSE & $\tau$ & $\rho$ & MSE & $\tau$ & $\rho$ & MSE & $\tau$ & $\rho$ & MSE \\
\midrule
\texttt{gpt-4o} & \textbf{.51} & \textbf{.56} & 2.03 & \textbf{.68} & \textbf{.77} & \textbf{0.72} & \textbf{.62} & \textbf{.71} & \textbf{0.97} & \textbf{.70} & \textbf{.78} & \textbf{0.81} & \textbf{0.49} & \textbf{0.53} & \textbf{0.63} \\
\rowcolor{gray!15} \texttt{~~w/ CoT} & .45 & .49 & \textbf{1.93} & .62 & .71 & 0.86 & .60 & .69 & 1.00 & .61 & .70 & 1.09 & 0.46 & 0.50 & 0.67 \\
\texttt{gpt-4.1-2025-04-14} & .52 & .56 & 1.61 & \textbf{.66} & \textbf{.75} & \textbf{0.78} & \textbf{.64} & \textbf{.73} & \textbf{0.84} & \textbf{.66} & \textbf{.76} & \textbf{0.84} & \textbf{.42} & \textbf{.46} & \textbf{0.66} \\
\rowcolor{gray!15} \texttt{~~w/ CoT} & \textbf{.60} & \textbf{.65} & \textbf{1.48} & .65 & .74 & 0.89 & .61 & .70 & 1.01 & .60 & .69 & 1.22 & .40 & .43 & 0.73 \\
\texttt{claude-3-5-sonnet} & .40 & .41 & 2.22 & \textbf{.60} & \textbf{.67} & \textbf{1.13} & \textbf{.59} & \textbf{.66} & \textbf{1.17} & .57 & .64 & \textbf{1.35} & 0.38 & 0.41 & \textbf{0.63} \\
\rowcolor{gray!15} \texttt{~~w/ CoT} & \textbf{.46} & \textbf{.48} & \textbf{2.07} & .54 & .62 & 1.17 & .56 & .64 & 1.47 & \textbf{.57} & \textbf{.64} & 1.53 & \textbf{0.51} & \textbf{0.54} & 0.71 \\
\texttt{claude-3-5-haiku} & .42 & .45 & \textbf{1.80} & \textbf{.54} & \textbf{.59} & 1.35 & \textbf{.56} & \textbf{.64} & \textbf{1.22} & \textbf{.55} & \textbf{.62} & \textbf{1.39} & \textbf{0.38} & \textbf{0.41} & 0.75 \\
\rowcolor{gray!15} \texttt{~~w/ CoT} & \textbf{.45} & \textbf{.48} & 1.86 & .54 & .61 & \textbf{1.21} & .47 & .54 & 1.58 & .35 & .39 & 1.85 & 0.39 & 0.42 & \textbf{0.66} \\
\texttt{gpt-4o-mini} & .27 & .31 & \textbf{2.74} & \textbf{.63} & \textbf{.70} & \textbf{0.93} & \textbf{.59} & \textbf{.67} & \textbf{0.93} & .60 & .67 & 1.08 & \textbf{0.37} & \textbf{0.40} & \textbf{0.63} \\
\rowcolor{gray!15} \texttt{~~w/ CoT} & \textbf{.29} & \textbf{.32} & 3.29 & .48 & .53 & 1.53 & .48 & .55 & 1.37 & \textbf{.60} & \textbf{.68} & \textbf{1.00} & 0.35 & 0.37 & \textbf{0.63} \\
\texttt{claude-3-haiku} & \textbf{.22} & \textbf{.23} & \textbf{2.17} & \textbf{.45} & \textbf{.50} & \textbf{1.55} & .42 & .47 & \textbf{1.47} & \textbf{.51} & \textbf{.56} & \textbf{1.51} & \textbf{0.20} & \textbf{0.21} & \textbf{0.81} \\
\rowcolor{gray!15} \texttt{~~w/ CoT} & .18 & .19 & 2.20 & .39 & .44 & 1.71 & \textbf{.43} & \textbf{.48} & 1.59 & .39 & .43 & 1.85 & 0.19 & 0.20 & 0.87 \\
\texttt{gemma-2-27b} & \textbf{.34} & \textbf{.38} & 2.08 & \textbf{.47} & \textbf{.52} & \textbf{1.27} & \textbf{.49} & \textbf{.56} & \textbf{1.24} & .49 & .55 & \textbf{1.25} & \textbf{0.18} & \textbf{0.19} & \textbf{0.70} \\
\rowcolor{gray!15} \texttt{~~w/ CoT} & .29 & .32 & \textbf{2.03} & .40 & .45 & 1.44 & .37 & .42 & 1.44 & \textbf{.52} & \textbf{.59} & 1.32 & 0.15 & 0.16 & 0.76 \\
\texttt{gemma-2-9b} & .28 & .30 & 2.36 & .43 & .48 & 1.40 & \textbf{.36} & \textbf{.41} & \textbf{1.47} & \textbf{.52} & \textbf{.58} & \textbf{1.34} & 0.09 & 0.09 & \textbf{0.73} \\
\rowcolor{gray!15} \texttt{~~w/ CoT} & \textbf{.31} & \textbf{.34} & \textbf{2.00} & \textbf{.45} & \textbf{.50} & \textbf{1.35} & .32 & .37 & 1.64 & .32 & .36 & 1.59 & \textbf{0.18} & \textbf{0.19} & 0.88 \\

\texttt{llama-3-8b} & -.01 & -.01 & \textbf{2.31} & \textbf{.37} & \textbf{.41} & 1.56 & \textbf{.29} & \textbf{.32} & \textbf{1.65} & .21 & .24 & 2.04 & 0.05 & 0.06 & 1.06 \\

\rowcolor{gray!15} \texttt{~~w/ CoT} & \textbf{.14} & \textbf{.14} & 2.44 & .15 & .16 & 1.91 & .21 & .24 & 1.80 & \textbf{.26} & \textbf{.29} & \textbf{1.81} & \textbf{0.15} & \textbf{0.16} & \textbf{0.92} \\

\midrule
\midrule
\texttt{deepseek-v3} & \textbf{.47} & \textbf{.52} & \textbf{1.75} & \textbf{.66} & \textbf{.74} & \textbf{0.90} & \textbf{.60} & \textbf{.67} & \textbf{0.96} & \textbf{.67} & \textbf{.74} & \textbf{0.95} & \textbf{.51} & \textbf{.55} & 0.75 \\
\rowcolor{gray!15} \texttt{~~w/ CoT} & .43 & .47 & 1.96 & .59 & .68 & 0.97 & .58 & .67 & 1.05 & .51 & .58 & 1.53 & .41 & .45 & 0.69 \\

\texttt{deepseek-r1} & \textbf{.59} & \textbf{.64} & \textbf{1.35} & \textbf{.57} & \textbf{.65} & \textbf{1.05} & \textbf{.57} & \textbf{.66} & \textbf{1.01} & \textbf{.59} & \textbf{.68} & \textbf{1.04} & \textbf{.50} & \textbf{.55} & \textbf{0.65} \\
\rowcolor{gray!15} \texttt{~~w/ CoT} & .52 & .57 & 1.60 & .54 & .63 & 1.12 & .55 & .63 & 1.11 & .55 & .65 & 1.15 & .38 & .41 & 0.63 \\

\texttt{o3-2025-04-16} & \textbf{.53} & \textbf{.57} & \textbf{1.40} & .65 & .73 & 0.87 & \textbf{.60} & \textbf{.68} & \textbf{0.93} & \textbf{.59} & \textbf{.67} & \textbf{1.31} & \textbf{.43} & \textbf{.47} & \textbf{0.63} \\
\rowcolor{gray!15} \texttt{~~w/ CoT} & .53 & .56 & 1.41 & \textbf{.68} & \textbf{.76} & \textbf{0.82} & .59 & .67 & 0.96 & .57 & .65 & 1.38 & .43 & .46 & 0.62 \\

\bottomrule
\end{tabular}
\caption{\small Effect of Chain of Thought across Step 1 RULER (Honorifics, Lexical, Syntax, Content) and Step 2 VERSE (Overall Aggregated). Results show that CoT has mixed effects: while most models benefit from CoT in evaluating Honorifics, for most models CoT does not improve performance, and in some cases even hurts. This pattern hold true even for the reasoning models where the models produce long reasoning trace before outputting answer. o3 was evaluated at $T=1.0$, the only available temperature setting.}
\label{tab:step1_2_results_cot}
\end{table*}

\begin{table*}[h]
\centering
\footnotesize
\captionsetup{font=small}
\setlength{\tabcolsep}{4pt}
\begin{tabular}{l|ccc|ccc|ccc|ccc|ccc}
\toprule
& \multicolumn{3}{c|}{\textbf{Honorifics}} & \multicolumn{3}{c|}{\textbf{Lexical}} & \multicolumn{3}{c|}{\textbf{Syntax}} & \multicolumn{3}{c|}{\textbf{Content}} & \multicolumn{3}{c}{\textbf{Step 2 (VERSE)}} \\
\cmidrule(lr){2-4} \cmidrule(lr){5-7} \cmidrule(lr){8-10} \cmidrule(lr){11-13} \cmidrule(lr){14-16}
\textbf{Eval Model} & $\tau$ & $\rho$ & MSE & $\tau$ & $\rho$ & MSE & $\tau$ & $\rho$ & MSE & $\tau$ & $\rho$ & MSE & $\tau$ & $\rho$ & MSE \\
\midrule
\texttt{gpt-4o} & \textbf{.51} & \textbf{.56} & 2.03 & \textbf{.68} & \textbf{.77} & \textbf{0.72} & .62 & .71 & .97 & \textbf{.70} & \textbf{.78} & \textbf{0.81} & 0.49 & 0.53 & 0.63 \\
\rowcolor{gray!15} ~~w/o rubric & .23 & .26 & 4.63 & \textbf{.68} & \textbf{.77} & 0.82 & .62 & .71 & \textbf{0.87} & .65 & .73 & 0.85 & 0.07 & 0.08 & 1.22 \\
\rowcolor{gray!15} ~~w/o human TR & .50 & .54 & \textbf{1.97} & .67 & .76 & 0.83 & \textbf{.63} & \textbf{.72} & 1.12 & .67 & .75 & 1.02 & \textbf{0.51} & \textbf{0.56} & \textbf{0.60} \\
\midrule
\texttt{gpt-4o-mini} & .27 & .31 & 2.74 & .63 & .70 & 0.93 & .59 & .67 & 0.93 & .60 & .67 & 1.08 & 0.37 & 0.40 & 0.63 \\
\rowcolor{gray!15} ~~w/o rubric & .20 & .22 & 3.34 & \textbf{.64} & \textbf{.71} & \textbf{0.91} & \textbf{.61} & \textbf{.68} & \textbf{0.95} & \textbf{.62} & \textbf{.69} & \textbf{0.97} & 0.03 & 0.04 & 1.02 \\
\rowcolor{gray!15} ~~w/o human TR & \textbf{.32} & \textbf{.35} & \textbf{2.55} & .59 & .66 & 0.96 & .56 & .63 & 1.02 & .59 & .66 & 1.15 & \textbf{0.40} & \textbf{0.44} & \textbf{0.62} \\
\midrule
\texttt{gemma-2-27b-it} & .34 & .38 & 2.08 & .47 & .52 & 1.27 & .49 & .56 & 1.24 & .49 & .55 & 1.25 & 0.18 & 0.19 & \textbf{0.70} \\
\rowcolor{gray!15} ~~w/o rubric & .32 & .33 & 3.96 & \textbf{.54} & \textbf{.60} & 1.48 & \textbf{.54} & \textbf{.61} & 1.17 & \textbf{.59} & \textbf{.66} & \textbf{1.09} & 0.10 & 0.11 & \textbf{0.70} \\
\rowcolor{gray!15} ~~w/o human TR & \textbf{.37} & \textbf{.41} & \textbf{1.91} & .52 & .58 & \textbf{1.12} & .52 & .59 & \textbf{1.10} & .54 & .61 & 1.17 & \textbf{0.24} & \textbf{0.25} & 0.75 \\
\midrule
\texttt{gemma-2-9b-it} & .28 & .30 & 2.36 & .43 & .48 & 1.40 & .36 & .41 & 1.47 & .52 & .58 & 1.34 & 0.09 & 0.09 & \textbf{0.73} \\
\rowcolor{gray!15} ~~w/o rubric & .20 & .21 & 3.88 & .30 & .34 & 1.83 & .29 & .33 & 1.73 & .24 & .27 & 1.74 & 0.00 & 0.00 & 0.83 \\
\rowcolor{gray!15} ~~w/o human TR & \textbf{.29} & \textbf{.32} & \textbf{2.11} & \textbf{.53} & \textbf{.59} & \textbf{1.36} & \textbf{.47} & \textbf{.53} & \textbf{1.52} & \textbf{.50} & \textbf{.56} & \textbf{1.52} & \textbf{0.16} & \textbf{0.17} & 0.76 \\
\midrule
\texttt{llama-3-8b-it} & -.01 & -.01 & \textbf{2.31} & \textbf{.37} & \textbf{.41} & \textbf{1.56} & \textbf{.29} & \textbf{.32} & \textbf{1.65} & \textbf{.21} & \textbf{.24} & \textbf{2.04} & \textbf{0.05} & \textbf{0.06} & 1.06 \\
\rowcolor{gray!15} ~~w/o rubric & -.05 & -.05 & 3.20 & .18 & .19 & 1.99 & .13 & .15 & 2.14 & .15 & .16 & 2.18 & -0.02 & -0.03 & \textbf{0.73} \\
\rowcolor{gray!15} ~~w/o human TR & \textbf{.11} & \textbf{.12} & 2.34 & .34 & .37 & 1.65 & \textbf{.29} & \textbf{.32} & 1.74 & .15 & .16 & 2.18 & 0.05 & 0.05 & 1.06 \\
\bottomrule
\end{tabular}
\caption{\small Ablation of the effect of human translation and rubric. Note that for most models, removing the rubric has a bigger effect than removing the human translation. In fact, for Step 2 VERSE, the models tend to do better \textit{without} human translation.}
\label{tab:(step 1_2) rubric and human tr ablation}
\end{table*}
\begin{figure*}[t]
  \centering
  \begin{subfigure}[b]{0.45\textwidth}
      \centering
      \includegraphics[width=0.95\textwidth]{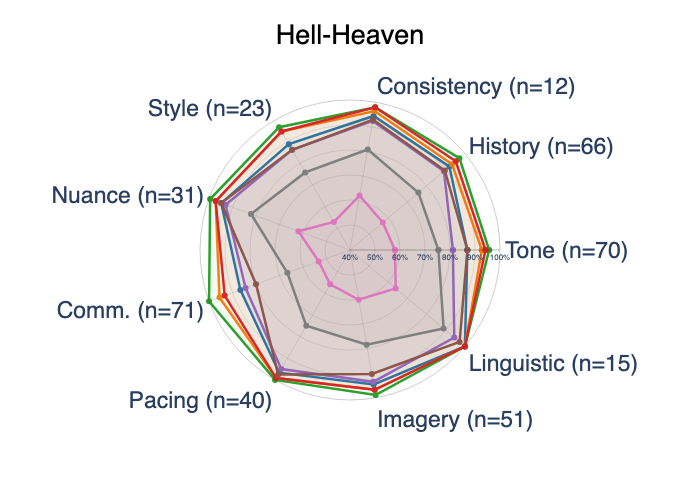}

      \label{fig:1}
  \end{subfigure}
  \hfill
  \begin{subfigure}[b]{0.45\textwidth}
      \centering
      \includegraphics[width=\textwidth]{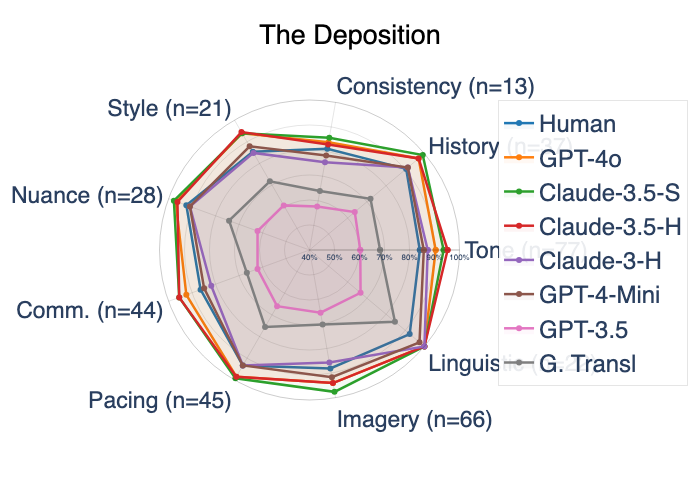}

      \label{fig:2}
  \end{subfigure}

  \vspace{1em}

  \begin{subfigure}[b]{0.45\textwidth}
      \centering
      \includegraphics[width=\textwidth]{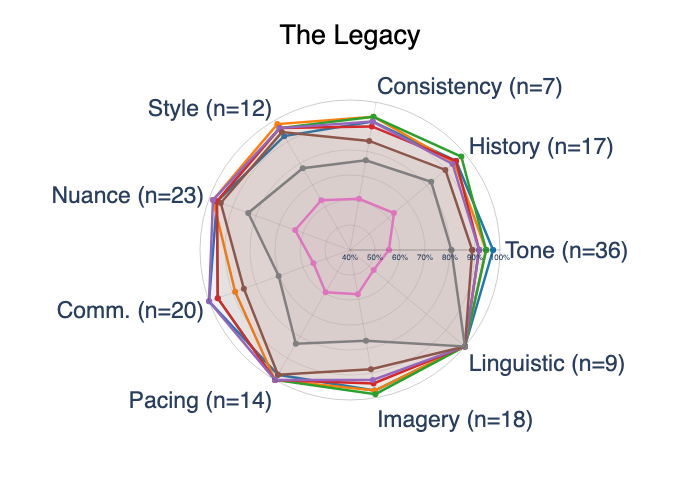}

      \label{fig:3}
  \end{subfigure}
  \hfill
  \begin{subfigure}[b]{0.45\textwidth}
      \centering
      \includegraphics[width=\textwidth]{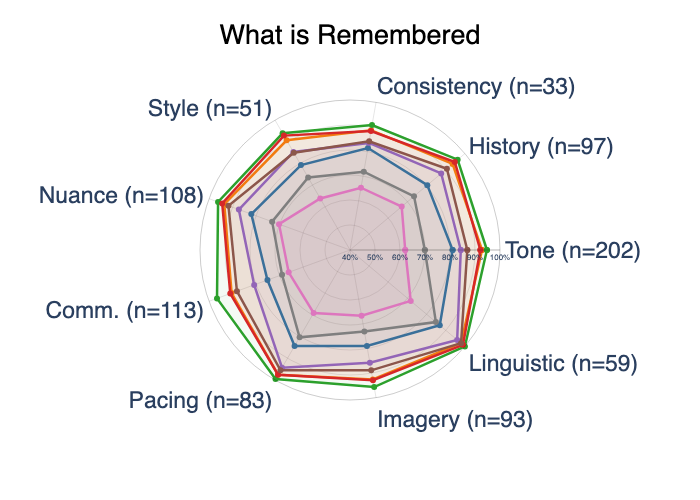}

      \label{fig:4}
  \end{subfigure}

  \vspace{1em}

  \begin{subfigure}[b]{0.45\textwidth}
      \centering
      \includegraphics[width=\textwidth]{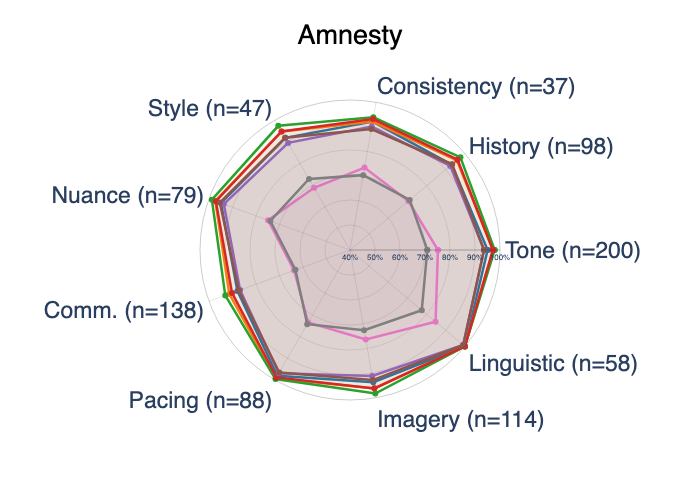}

      \label{fig:5}
  \end{subfigure}
  \hfill
  \begin{subfigure}[b]{0.45\textwidth}
      \centering
      \includegraphics[width=\textwidth]{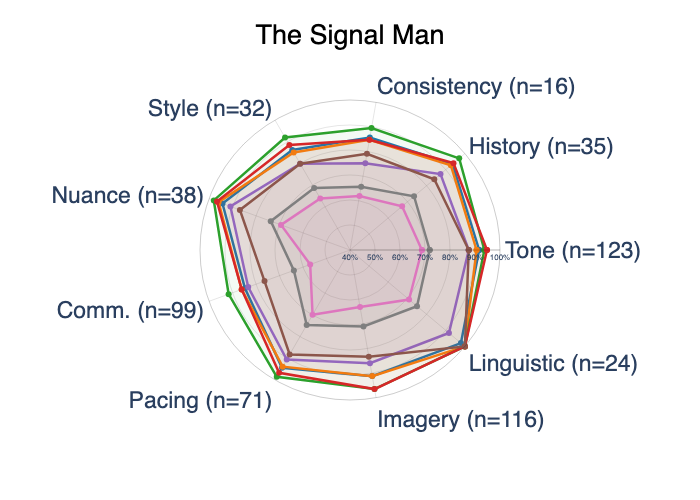}

      \label{fig:6}
  \end{subfigure}



  \vspace{1em}

  \begin{subfigure}[b]{0.45\textwidth}
      \centering
      \includegraphics[width=0.95\textwidth]{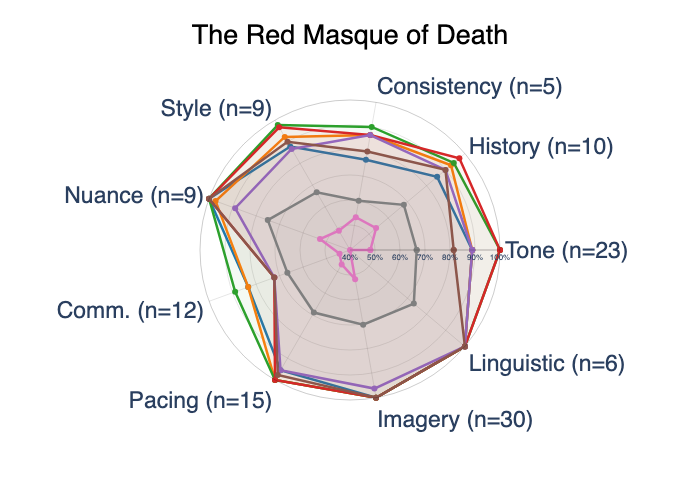}

      \label{fig:9}
  \end{subfigure}
  \hfill
  \begin{subfigure}[b]{0.45\textwidth}
      \centering
      \includegraphics[width=0.95\textwidth]{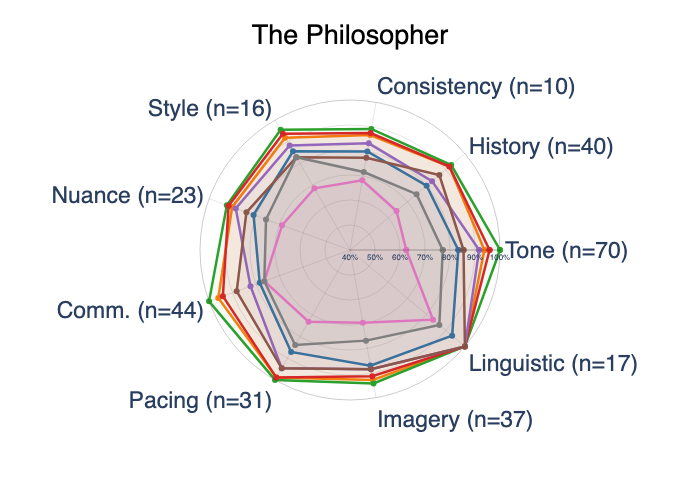}

      \label{fig:10}
  \end{subfigure}
  \caption{\small Fine-grained evaluation of 8 stories using VERSE. Notice that models score widely differently for different stories. The difference is more pronounced for less powerful model, for whom evaluator LLM is more sensitive. $n$ refers to the number of questions belonging to that category. The legend on the bottom left shows which color belongs to which model.}
  \label{fig:step 2 all stories evaluated}
\end{figure*}

\end{document}